\documentclass{article} 
\usepackage{iclr2027_conference,times}
\iclrfinalcopy

\usepackage{amsmath,amsfonts,bm}

\def\eqref#1{equation~\ref{#1}}

\def\1{\bm{1}}

\DeclareMathAlphabet{\mathsfit}{\encodingdefault}{\sfdefault}{m}{sl}
\SetMathAlphabet{\mathsfit}{bold}{\encodingdefault}{\sfdefault}{bx}{n}

\usepackage{graphicx} \graphicspath{{figures/}}
\usepackage{amsmath,amssymb,mathabx,mathtools,amsthm,nicefrac}
\usepackage{subcaption}
\usepackage{caption}
\usepackage{acronym}
\usepackage{wrapfig}
\usepackage[dvipsnames,svgnames,x11names,table]{xcolor}
\usepackage{enumerate}
\usepackage{xspace}
\usepackage[misc]{ifsym}
\usepackage{titletoc}
\usepackage{verbatim,fancyvrb,listings}
\usepackage{booktabs,tabularx,colortbl,multirow,multicol,array,makecell,tabularray}

\usepackage[pagebackref,breaklinks,colorlinks]{hyperref}
\usepackage[capitalise,noabbrev]{cleveref}

\makeatletter
\DeclareRobustCommand\onedot{\futurelet\@let@token\@onedot}
\def\@onedot{\ifx\@let@token.\else.\null\fi\xspace}

\makeatother

\definecolor{cardinal}{rgb}{0.549,0.082,0.082}
\acrodef{2afc}[2AFC]{two-alternative forced-choice}
\acrodef{dpo}[DPO]{direct preference optimization}
\acrodef{mlp}[MLP]{multilayer perceptron}
\acrodef{mpjpe}[MPJPE]{mean per-joint position error}
\acrodef{pampjpe}[PA-MPJPE]{Procrustes-aligned MPJPE}
\acrodef{ppo}[PPO]{proximal policy optimization}
\acrodef{rte}[RTE]{root translation error}
\acrodef{sft}[SFT]{supervised fine-tuning}
\newcommand{\model}{\textcolor{cardinal}{\textbf{ViT-H}}\xspace}
\newcommand{\modelname}{\textcolor{cardinal}{ViT-H}\xspace}

\addtocontents{toc}{\protect\setcounter{tocdepth}{-1}}

\title{\modelname: Generative \textcolor{cardinal}{\underline{Vi}}deo-\textcolor{cardinal}{\underline{T}}o\textcolor{cardinal}{\underline{-H}}umanoid\\Control with Motion Tracking Feedback}

\author{
    \setlength{\tabcolsep}{3pt}
    \begin{tabular}{llll}
    Zimo He \textsuperscript{1, 2, 4, $\star$} &
    Tianyi Bian \textsuperscript{3, 4, 5, $\star$} &
    Zizhong Yang \textsuperscript{3, 4, 5, $\star$} &
    Hongjie Li \textsuperscript{3, 4, $\star$, $\dagger$} \\
    Nan Jiang \textsuperscript{4, 6, 7, 8} &
    Jieming Cui \textsuperscript{3, 6, 7, 8} &
    Baoxiong Jia \textsuperscript{2, 6} & \\
    Yizhou Wang \textsuperscript{1, 6, $\textrm{\Letter}$} &
    Siyuan Huang \textsuperscript{2, 6, $\textrm{\Letter}$} &
    Yixin Zhu \textsuperscript{4, 3, 6, 7, 8, 9, $\textrm{\Letter}$}
    \end{tabular}\vspace{3pt}\\
    \small$^\star$\,Equal contributions \quad $^\dagger$\,Project lead \quad $^\textrm{\Letter}$\,Corresponding authors \vspace{4pt}\\
    \small\textsuperscript{1} School of Computer Science, Peking University \quad 
    \small\textsuperscript{2} Beijing Institute for General Artificial Intelligence \\
    \small\textsuperscript{3} Institute for AI, Peking University \quad
    \small\textsuperscript{4} School of Psychological and Cognitive Sciences, Peking University \\
    \small\textsuperscript{5} Yuanpei College, Peking University \quad
    \small\textsuperscript{6} State Key Lab of General AI \\
    \small\textsuperscript{7} Beijing Key Laboratory of Brain-Computer Interface and Mental Health Modulation \\
    \small\textsuperscript{8} Embodied Intelligence Lab, PKU-Wuhan Institute for AI \quad
    \small\textsuperscript{9} XS Vision
}

\begin{document}
\maketitle
\lhead{Preprint. Under review.}

\begin{center}
    \vspace{-2em}
    \includegraphics[width=\linewidth]{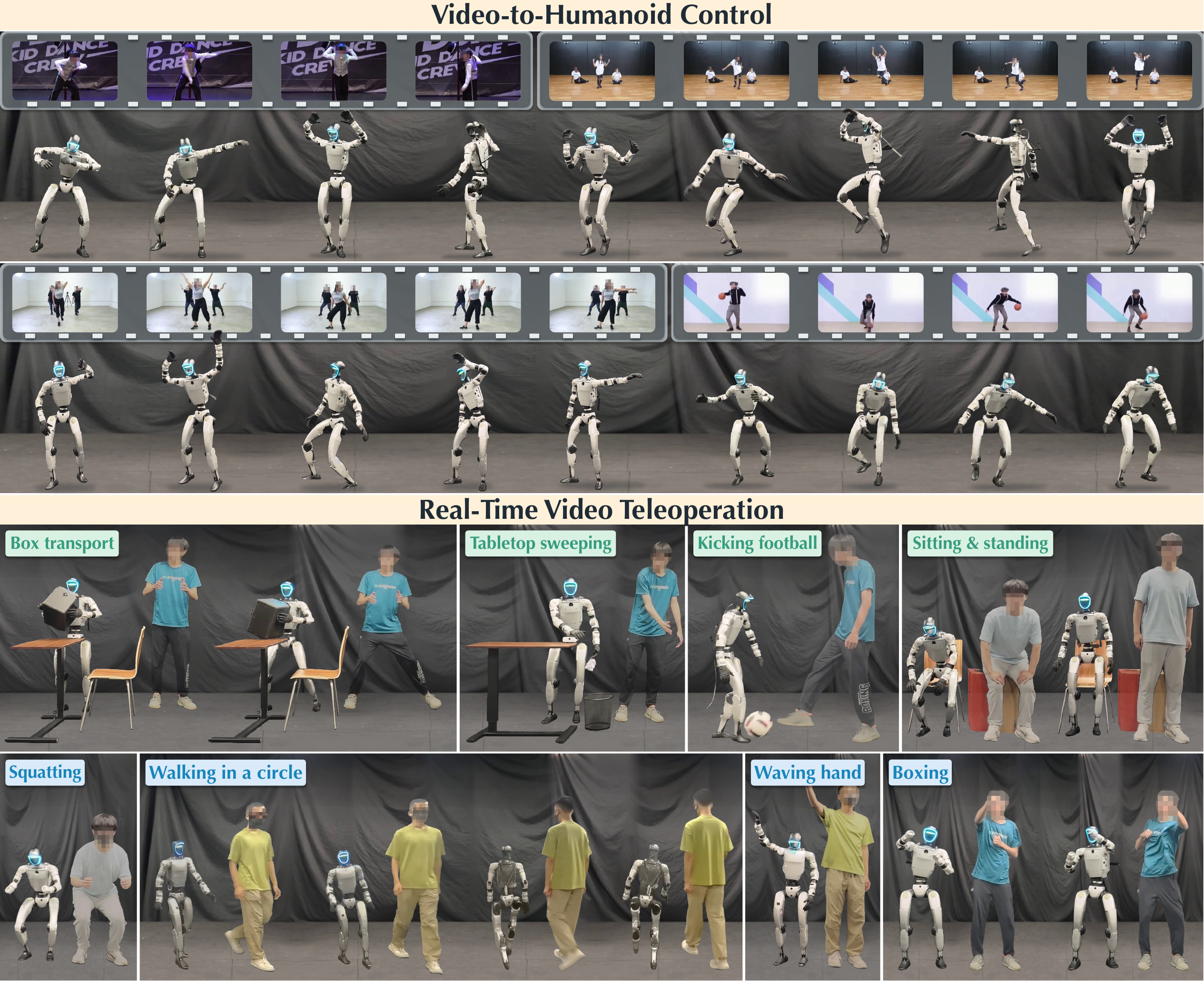}
    \captionof{figure}{\textbf{\model transfers human motions from monocular videos to a physical Unitree G1.} The top rows show offline video-to-humanoid control. Each film strip shows frames of an input video, and the frames below show the robot tracking the reference motion generated for the whole video at once. The bottom rows show real-time teleoperation from a single RGB camera, in which the autoregressive generator produces reference motions segment by segment for tasks spanning locomotion and loco-manipulation. Please refer to our \href{https://awfuact.github.io/vit-h/}{\textit{project page}} for video results.}
    \label{fig:teaser}
\end{center}

\begin{abstract}
    Learning humanoid control from monocular human videos offers a scalable way to acquire diverse whole-body behaviors.
    Yet, existing reconstruction-retargeting-tracking pipelines 
    are limited in system-level scalability by separately optimized intermediate stages and cannot directly optimize reference motion executability using downstream tracking feedback.
    We present \model, a video-to-humanoid control framework following a generation-and-tracking paradigm.
    \model directly generates humanoid reference motions from videos with a diffusion model, bypassing human motion reconstruction and retargeting, and uses motion-tracking outcomes to post-train the generator with direct preference optimization. 
    We further introduce an autoregressive variant that enables real-time humanoid video teleoperation with sub-400 ms latency.
    Extensive experiments on the Unitree G1 demonstrate that \model outperforms state-of-the-art reconstruction-retargeting-tracking pipelines in both motion quality and executability.
    We further show that \model scales favorably with model size and with both pre-training and post-training data scale, while motion-tracking feedback consistently improves executability across different motion trackers. Project page: \href{https://awfuact.github.io/vit-h/}{\textit{https://awfuact.github.io/vit-h/}}.
\end{abstract}

\acresetall
\section{Introduction}

Monocular human videos record whole-body behaviors at a scale and diversity that motion capture datasets~\citep{mahmood2019amass} cannot match, making them an attractive source of skills for humanoid robots. Recent works have leveraged human videos to learn expressive whole-body skills~\citep{he2025asap,xie2025kungfubot,sleiman2026zest}, environment-aware behaviors that interact with surrounding geometry~\citep{allshire2025videomimic}, and interactive skills involving objects and humans~\citep{wang2026humanx}. Most existing systems realize this capability through a reconstruction-retargeting-tracking pipeline, which first reconstructs human motion from video, then retargets it to the humanoid embodiment, and finally tracks the resulting reference motion. This modular separation is increasingly practical as recent advances in general whole-body tracking enable a shared tracker to execute diverse reference motions~\citep{luo2026sonic,zeng2026scaling}. However, the upstream reconstruction-and-retargeting stages remain an indirect route from video to humanoid motion, where errors can accumulate to produce inconsistent, distorted, or dynamically infeasible references, making downstream tracking more difficult.

Prior work has sought to improve reference motion quality by strengthening individual stages of this pipeline, either by enforcing physical plausibility in reconstructed human motions~\citep{zhang2024physpt,zhang2025plug,feng2025physhmr} or by improving motion retargeting~\citep{araujo2025retargeting,lee2026phuma,zhao2026make,cao2026unified}. However, such stage-local remedies expose two structural limitations of the pipeline. First, its \textbf{system-level scalability remains bounded by the remaining stages}. Human motion reconstruction is not optimized for the morphology and dynamics of the target humanoid, while retargeting can only operate on reconstructed motions after upstream errors have propagated and discarded visual evidence can no longer be revisited. Second, \textbf{executability is difficult to optimize end-to-end}. Tracking results are observed only after reconstruction and retargeting, making it difficult to use execution outcomes to directly improve the complete video-to-humanoid mapping. Consequently, plausible reference motions at intermediate stages may still be difficult for the control policy to track.

To this end, we introduce \model, a generative \textcolor{cardinal}{\textbf{\underline{Vi}}}deo-\textcolor{cardinal}{\textbf{\underline{T}}}o\textcolor{cardinal}{\textbf{\underline{-H}}}umanoid control framework following a generation-and-tracking paradigm. At the core of \model are (i) a motion generator that maps monocular videos directly into the target humanoid motion space, eliminating the intermediate bottleneck of human motion reconstruction and retargeting, and (ii) whole-body control policies that not only track the generated reference motions but also evaluate their executability, providing motion-tracking feedback for post-training the reference motion generator, thereby closing the loop between generation and control (\cref{fig:method}).

Concretely, our motion generator is a video-conditioned humanoid motion diffusion model~\citep{ho2020denoising} first pre-trained on video-motion pairs and motion-only data. The whole-body control policies are fine-tuned from a pre-existing general motion tracker, yielding a deployable tracking policy and a privileged variant with additional global observations. We evaluate motions sampled from the generator with both tracking policies and use the resulting tracking outcomes to construct preference pairs for \ac{dpo}~\citep{rafailov2023direct,wallace2024diffusion}, explicitly incorporating executability into the humanoid motion generator. Because preference pair construction needs no ground-truth motion, post-training extends to unannotated internet videos. Additionally, we further extend \model with an autoregressive variant of the motion generator that produces motions in temporal segments, enabling online control from streaming video and real-time video teleoperation of a physical Unitree G1 from a single RGB camera (\cref{fig:teaser}).

Our contributions are fourfold. First, we introduce \model, a generative video-to-humanoid control framework that directly generates executable humanoid motions from monocular videos without human motion reconstruction and retargeting. Second, we introduce \ac{dpo} post-training with motion-tracking feedback, which explicitly incorporates executability into reference motion generation while requiring no motion annotations. Third, extensive experiments demonstrate that \model outperforms state-of-the-art reconstruction-retargeting-tracking baselines in motion quality and executability. Fourth, we demonstrate that \model scales favorably with model capacity and both pre-training and post-training data size, while its post-training gains generalize to three held-out general motion trackers~\citep{qi2026humanoid,zeng2026scaling,chen2026holomotion}.

\section{Related Work}

\paragraph{Human motion reconstruction from video.} 
Monocular 3D human reconstruction has been extensively studied with parametric body models~\citep{loper2015smpl,pavlakos2019expressive,saito2026soma}. Early approaches primarily estimate human pose and shape from individual images~\citep{kanazawa2018end,kolotouros2019learning,kocabas2021pare,li2021hybrik,zhang2023pymaf}, while video-based methods exploit temporal information to improve motion consistency and robustness~\citep{kocabas2019vibe,wan2021encoder,shen2023global}. Recent work extends human motion recovery beyond camera coordinates by estimating globally consistent trajectories and disentangling human and camera motion~\citep{yuan2022glamr,ye2023decoupling,sun2023trace,kocabas2024pace,shin2024wham,wang2024tram,shen2024world,li2025genmo,zhao2026onlinehmr}. Complementary approaches improve the physical plausibility of reconstructed human motions through dynamics-aware refinement, motion restoration, or physics-based character control~\citep{zhang2024physpt,zhang2025plug,feng2025physhmr}. While these methods operate in the human motion space, our goal is not to recover physically plausible human motion as an intermediate representation, but to directly generate robot-space humanoid reference motions and optimize them for downstream tracking executability.

\paragraph{Human-to-humanoid motion retargeting.} 
Motion retargeting converts human motions into humanoid robot joint space through geometric alignment and optimization-based inverse kinematics~\citep{he2024learning,he2024omnih2o}. GMR~\citep{araujo2025retargeting} improves general-purpose retargeting by reducing kinematic artifacts and preserving source-motion fidelity, while PHUMA~\citep{lee2026phuma} introduces physical constraints to enforce joint limits and contact consistency. Concurrent work NMR~\citep{zhao2026make} learns a direct human-to-humanoid motion mapping to facilitate downstream whole-body tracking. In contrast, our approach bypasses human-to-humanoid retargeting during inference and directly generates robot-specific reference motions from monocular videos.

\paragraph{Humanoid motion generation and whole-body tracking.} 
Whole-body motion tracking provides a scalable interface for executing diverse kinematic reference motions on humanoid robots~\citep{ze2025twist,chen2025gmt,liao2026beyondmimic,luo2026sonic,qi2026humanoid,zeng2026scaling,chen2026holomotion,chen2026scenebot}. Building on such trackers, recent works generate text-conditioned humanoid motion references for whole-body control~\citep{jiang2024harmon,rempe2026kimodo,zhang2026learning,harithas2026dreamcontrol}, extend motion generation to multimodal conditioning~\citep{jiang2025uniact,huang2026omg}, and predict kinematic trajectories for loco-manipulation~\citep{wang2026vlk}. Our approach follows this generation-and-tracking paradigm, but rather than optimizing the two components independently, we use tracking feedback to post-train the generator for executability.

Another line of work further collapses explicit motion generation and tracking, enabling retargeting-free humanoid control from language~\citep{li2026language}, audio~\citep{li2026do}, or video~\citep{li2026robomirror} by training dedicated control policies directly conditioned on motion latents. While this formulation eliminates explicit motion decoding, it requires a specialized control policy for the latent interface and is therefore less able to directly leverage advances in general motion trackers.

\paragraph{Post-training of motion generation models.}
Originally developed for language models, \ac{dpo}~\citep{rafailov2023direct} has been extended to diffusion models~\citep{wallace2024diffusion} and applied to human motion generation to improve motion quality and text alignment~\citep{pappa2026modipo,tan2025sopo}. Closer to our setting, recent work post-trains text-conditioned human motion generators with feedback from physics-based tracking or a whole-body controller to improve physical plausibility~\citep{yue2025rl,zhang2026physmodpo}. Despite enabling deployment on humanoid robots, these methods require motion retargeting and optimize motions in the human motion space rather than for the target humanoid embodiment. Our approach instead post-trains the video-conditioned motion generator directly in the target humanoid motion space.

\section{\modelname}

\paragraph{Formulation.}
Our goal is to convert the human motion in a monocular video $\mathbf{I}=\{\mathbf{I}_t\}_{t=1}^T$ into whole-body actions that make a Unitree G1 humanoid reproduce it. We decompose this problem into reference motion generation and whole-body tracking. A video-conditioned generative model $\pi_\text{gen}$ predicts a reference humanoid motion $\mathbf{X}\sim\pi_\text{gen}(\cdot\mid\mathbf{I})$, and a whole-body tracking policy $\pi_\text{track}$ executes it. At each step, $\pi_\text{track}$ outputs an action $\mathbf{a}_t\sim\pi_\text{track}(\cdot\mid\mathbf{s}_t,\mathbf{X}_t)$ from the robot state $\mathbf{s}_t$ and the current and upcoming reference frames $\mathbf{X}_t$, and a proportional-derivative controller converts $\mathbf{a}_t$ into joint torques. We consider \textit{offline} generation, which produces the complete reference motion at once, and \textit{online} generation, which produces it autoregressively in temporal segments conditioned on the preceding motion.

\begin{figure*}[t!]
    \centering
    \includegraphics[width=\linewidth]{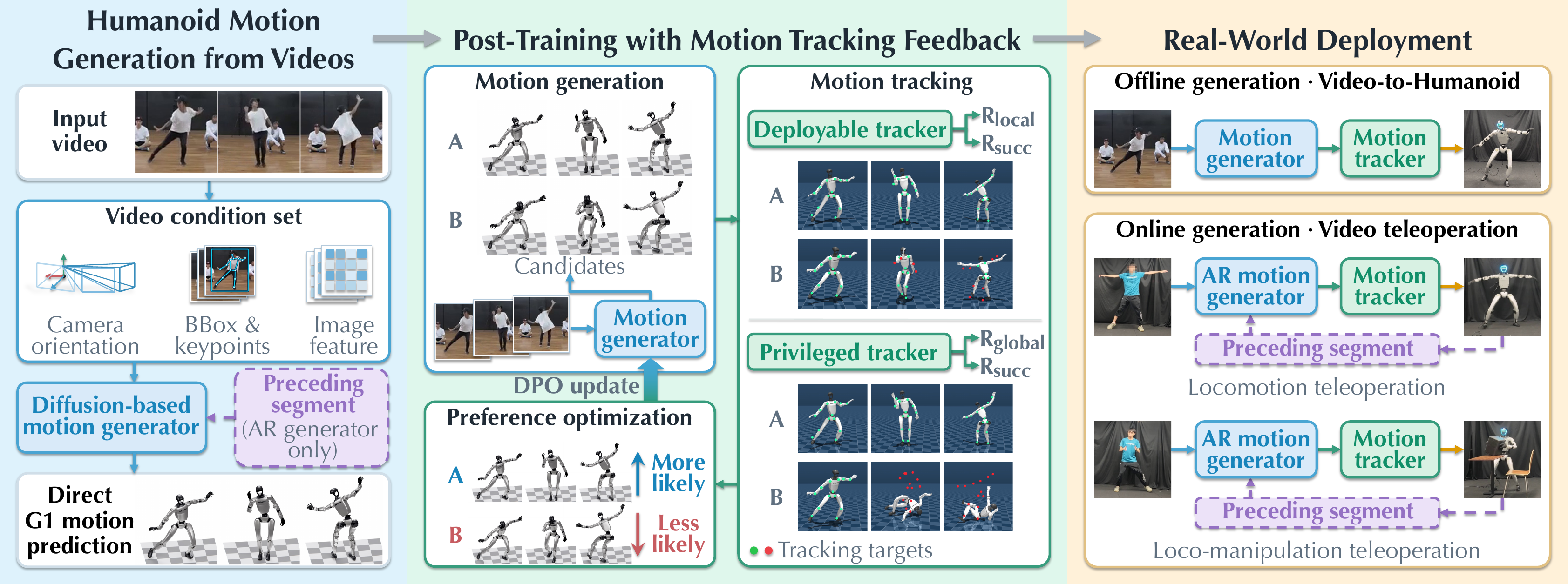}
    \caption{\textbf{Overview of \model.} The diffusion-based generator (left) maps the camera orientations, bounding boxes, keypoints, and image features of a video directly to G1 reference motions, and its autoregressive variant also conditions on the preceding segment. For post-training (middle), candidates sampled for the same video are executed by a deployable tracker, which scores local pose accuracy and success, and by a privileged tracker, which scores global motion accuracy and success. At deployment (right), the offline generator drives video-to-humanoid control and the autoregressive generator drives real-time teleoperation.}
    \label{fig:method}
\end{figure*}

\paragraph{Overview.} 
We propose \model, a generation-and-tracking framework for video-driven humanoid control. An overview of \model is shown in \cref{fig:method}. We first train a diffusion-based motion generator $\pi_\text{gen}$ to directly predict reference humanoid motions from video, eliminating the need for human motion reconstruction and retargeting at inference (\cref{sec:motion-generation}). We then fine-tune a general motion tracker~\citep{luo2026sonic} into a deployable tracking policy $\pi_\text{track}$ and a privileged variant $\pi_\text{track}^\text{priv}$ with additional global observations. Beyond executing reference motions, the two trackers score motions sampled from $\pi_\text{gen}$, and these scores construct preference pairs for post-training $\pi_\text{gen}$ with \ac{dpo} (\cref{sec:post-train}). For the online setting, an autoregressive variant $\pi_\text{gen}^\text{ar}$ supports real-time video teleoperation (\cref{sec:autoregressive-generation}). \Cref{sec:deploy-detail} details real-world deployment.

\subsection{Humanoid Motion Generation from Videos}\label{sec:motion-generation}

\paragraph{Data representation.} 
We represent video frames with condition set $\mathcal{C}=\{\mathbf{c}_\text{cam},\mathbf{c}_\text{bbox},\mathbf{c}_\text{kp2d},\mathbf{c}_\text{feat}\}$, where $\mathbf{c}_\text{cam}\in\mathbb{R}^{T\times6}$ denotes relative camera orientations, $\mathbf{c}_\text{bbox}\in\mathbb{R}^{T\times3}$ denotes 2D bounding boxes, $\mathbf{c}_\text{kp2d}\in\mathbb{R}^{T\times2K}$ denotes 2D keypoints, and $\mathbf{c}_\text{feat}\in\mathbb{R}^{T\times1024}$ denotes image features.

Inspired by the gravity-view coordinate system~\citep{shen2024world} introduced for human motion reconstruction, we represent the humanoid motion $\mathbf{X}$ with: (i) root velocity $\mathbf{v}_\text{root}\in\mathbb{R}^{T\times3}$, camera-view root orientation $\boldsymbol{\Gamma}_\text{cam}\in\mathbb{R}^{T\times6}$, and gravity-view root orientation $\boldsymbol{\Gamma}_\text{gv}\in\mathbb{R}^{T\times6}$ to parameterize the global trajectory, and (ii) local joint rotations $\mathbf{Q}\in\mathbb{R}^{T\times J\times6}$ encoded using the 6D representation~\citep{zhou2019continuity}. Additionally, we predict the camera-relative scale and translation $\boldsymbol{\tau}\in\mathbb{R}^{T\times3}$ and end-effector stationary probabilities $\mathbf{p}_\text{ee}\in\mathbb{R}^{T\times6}$, resulting in the full humanoid motion $\mathbf{X}=(\mathbf{v}_\text{root},\boldsymbol{\Gamma}_\text{cam},\boldsymbol{\Gamma}_\text{gv},\mathbf{Q},\boldsymbol{\tau},\mathbf{p}_\text{ee})$. Details on data representation are in \cref{sec:data-representation-detail}.

\paragraph{Video-conditioned humanoid motion generation.} 
The reference humanoid motion generator $\pi_\text{gen}$ is a diffusion model that takes the video-derived condition set $\mathcal{C}$ as conditional input. Following the DDPM framework~\citep{ho2020denoising}, we implement a forward diffusion process that progressively corrupts clean humanoid motions with Gaussian noise over $N$ steps:
\begin{equation}
q(\mathbf{X}_n\vert\mathbf{X}_{n-1})=\mathcal{N}(\mathbf{X}_n;\sqrt{1-\beta_n}\mathbf{X}_{n-1},\beta_n\bm{I}),
\end{equation}
where $n$ denotes the diffusion step and $\beta_n$ specifies the variance schedule. The generator $\pi_\text{gen}$ learns the reverse process, iteratively denoising samples initialized from $\mathcal{N}(\mathbf{0},\bm{I})$. We adopt the $\mathbf{X}_0$-prediction parameterization and train $\pi_\text{gen}$ to recover the clean motion from noisy samples directly:
\begin{equation}\label{eq:diffusion-loss}
    \mathcal{L}_\text{diffusion}=\mathbb{E}_{n\sim\mathcal{U}\{1,\ldots,N\},\mathbf{X}_n\sim q(\mathbf{X}_n\vert\mathbf{X}_0)}\big\Vert\mathbf{X}_0-\pi_\text{gen}(\mathbf{X}_n,n,\mathcal{C})\big\Vert_2^2.
\end{equation}
Additionally, we recover the global trajectory from the predicted $\hat{\mathbf{X}}_0$ and apply forward kinematics to obtain joint and vertex positions, which are used to compute losses on joints, vertices, root translation, end-effector stationarity, and 2D reprojection. Details on network architecture and losses are in \cref{sec:network-architecture-detail,sec:motion-generator-detail}, respectively.

\subsection{Post-Training with Motion Tracking Feedback}\label{sec:post-train}

\paragraph{Humanoid motion tracking.}
Our tracking policy $\pi_\text{track}$ is fine-tuned from SONIC~\citep{luo2026sonic}, a general humanoid motion tracker trained with \acl{ppo} \citep[PPO;][]{schulman2017proximal}\acused{ppo}. To make it follow the global trajectory of the reference, we add rewards for tracking the horizontal position and yaw of the anchor body, and give the critic the current and future root transformations of both the robot and the reference motion.

The actor of $\pi_\text{track}$ does not observe the root pose offset between the robot and the reference, so global deviations in its rollouts mix errors in the predicted trajectory with the tracker's own drift. We therefore fine-tune a privileged variant $\pi_\text{track}^\text{priv}$ from $\pi_\text{track}$ whose actor also observes this offset, keeping the remaining setup unchanged. \Cref{sec:motion-tracker-detail} details the training of both trackers.

\paragraph{Preference pair construction.} 
For each video-derived condition set $\mathcal{C}$, we sample $M$ candidates ${\{\mathbf{X}^{(i)}}\}_{i=1}^M \sim \pi_\text{gen}(\cdot \mid \mathcal{C})$ using our motion generator. We denote by $\bar{\mathbf{X}}^{(i)}$ and $\bar{\mathbf{X}}_\text{priv}^{(i)}$ the corresponding rollout motions obtained with $\pi_\text{track}$ and $\pi_\text{track}^\text{priv}$, respectively. We score each candidate with a local pose reward, a global motion reward, and a tracking-success reward. Let $\mathcal{E}_{\text{root-pos}}$, $\mathcal{E}_{\text{root-ori}}$, $\mathcal{E}_{\text{pos}}$, and $\mathcal{E}_{\text{ori}}$ denote the root translation, root orientation, root-relative joint position, and joint rotation errors, respectively. We first convert each error into a normalized reward 
\begin{equation}\label{eq:normalized-error}
    \rho_k(\mathbf{X},\mathbf{X}') = \exp\big(-\mathcal{E}_k(\mathbf{X},\mathbf{X}')/s_k^2\big)\in(0,1], 
\end{equation} 
where $s_k$ denotes the scale for error type $k$. The local and global rewards are then defined as 
\begin{align} 
    \mathcal{R}_{\text{local}}(\mathbf{X},\mathbf{X}') &= \boldsymbol{\alpha}_\text{local}^{\top}\big[\rho_{\text{pos}}(\mathbf{X},\mathbf{X}'), \rho_{\text{ori}}(\mathbf{X},\mathbf{X}')\big]^{\top}, \label{eq:reward-local}\\ 
    \mathcal{R}_{\text{global}}(\mathbf{X},\mathbf{X}') &= \boldsymbol{\alpha}_\text{global}^{\top}\big[\rho_{\text{root-pos}}(\mathbf{X},\mathbf{X}'),\rho_{\text{root-ori}}(\mathbf{X},\mathbf{X}'),\rho_{\text{pos}}(\mathbf{X},\mathbf{X}'),\rho_{\text{ori}}(\mathbf{X},\mathbf{X}')\big]^{\top}, \label{eq:reward-global}
\end{align} 
where $\boldsymbol{\alpha}_{\text{local}}$ and $\boldsymbol{\alpha}_{\text{global}}$ are normalized weights. We define the tracking-success reward as 
\begin{equation}\label{eq:reward-succ}
    \mathcal{R}_{\text{succ}}(s,s_{\text{priv}}) = \mathbb{I}[s=1 \land s_{\text{priv}}=1], 
\end{equation} 
where $s$ and $s_{\text{priv}}$ indicate whether the corresponding trackers successfully complete the rollout.

Using these formulations, we consider three rewards that do not require ground-truth motion: 
\begin{equation}\label{eq:reward-wo-gt}
    r_1^{(i)} = \mathcal{R}_{\text{local}} \big(\mathbf{X}^{(i)},\bar{\mathbf{X}}^{(i)}\big), \quad r_2^{(i)} = \mathcal{R}_{\text{global}} \big(\mathbf{X}^{(i)},\bar{\mathbf{X}}_{\text{priv}}^{(i)}\big), \quad r_3^{(i)} = \mathcal{R}_{\text{succ}} \big(s^{(i)},s_{\text{priv}}^{(i)}\big). 
\end{equation} 
When ground-truth humanoid motion $\mathbf{X}_{\text{GT}}$ is available, we additionally consider \begin{equation} 
    r_4^{(i)} = \mathcal{R}_{\text{global}} \big(\mathbf{X}^{(i)},\mathbf{X}_{\text{GT}}\big), \quad r_5^{(i)} = \mathcal{R}_{\text{local}} \big(\mathbf{X}_{\text{GT}},\bar{\mathbf{X}}^{(i)}\big), \quad r_6^{(i)} = \mathcal{R}_{\text{global}} \big(\mathbf{X}_{\text{GT}},\bar{\mathbf{X}}_{\text{priv}}^{(i)}\big). 
\end{equation}

We form a preference pair $(\mathbf{X}^{(i)}, \mathbf{X}^{(j)})$ when the reward advantage of $\mathbf{X}^{(i)}$ over $\mathbf{X}^{(j)}$ exceeds a margin $\boldsymbol{\delta}$ along each reward dimension: $\mathbf{r}^{(i)} - \mathbf{r}^{(j)} \succeq \boldsymbol{\delta}$. Additionally, a sample is eligible to serve as the preferred sample only if it is successfully tracked by both trackers. Let $\mathcal{P}$ denote the set of all candidate pairs satisfying this condition. We select the pair with the largest total reward gap:
\begin{equation}\label{eq:pair-select}
    (i^\star, j^\star) = \arg\max_{(i,j)\in\mathcal{P}}\mathbf{w}^{\top}\big(\mathbf{r}^{(i)} - \mathbf{r}^{(j)}\big),
\end{equation}
where $\mathbf{w}$ denotes the weights. We finally construct the pair $(\mathbf{X}^{+}, \mathbf{X}^{-}; \mathcal{C}) = (\mathbf{X}^{(i^\star)}, \mathbf{X}^{(j^\star)}; \mathcal{C})$.

\paragraph{\texorpdfstring{\ac{dpo}}{DPO} post-training.}
We fine-tune $\pi_\text{gen}$ against a frozen copy of its pre-trained parameters, $\pi_\text{gen}^\text{ref}$, which serves as the reference model. For each pair, we sample a diffusion step $n$ and a shared Gaussian noise $\boldsymbol{\epsilon}$, which yield the noisy motions $\mathbf{X}_n^{+}\sim q(\cdot\mid\mathbf{X}^{+})$ and $\mathbf{X}_n^{-}\sim q(\cdot\mid\mathbf{X}^{-})$. Under the $\mathbf{X}_0$-prediction parameterization, the per-sample denoising loss of a model $\pi$ is
\begin{equation}
    \ell_{\pi}(\mathbf{X},\mathbf{X}_n,n,\mathcal{C})=\big\Vert\mathbf{X}-\pi(\mathbf{X}_n,n,\mathcal{C})\big\Vert_2^2.
\end{equation}
Following Diffusion-DPO~\citep{wallace2024diffusion}, the \ac{dpo} objective is
\begin{equation}\label{eq:dpo}
    \begin{split}
        \mathcal{L}_\text{DPO} = -\mathbb{E}\Big[\log\sigma\Big(&\beta_\text{DPO}\big[\ell_{\pi_{\text{gen}}}(\mathbf{X}^{-},\mathbf{X}_n^{-},n,\mathcal{C})-\ell_{\pi_{\text{gen}}^{\text{ref}}}(\mathbf{X}^{-},\mathbf{X}_n^{-},n,\mathcal{C})\big]\\
        -&\beta_\text{DPO}\big[\ell_{\pi_{\text{gen}}}(\mathbf{X}^{+},\mathbf{X}_n^{+},n,\mathcal{C})-\ell_{\pi_{\text{gen}}^{\text{ref}}}(\mathbf{X}^{+},\mathbf{X}_n^{+},n,\mathcal{C})\big]\Big)\Big],
    \end{split}
\end{equation}
where $\sigma$ is the sigmoid function and $\beta_\text{DPO}$ controls how far $\pi_\text{gen}$ may deviate from $\pi_\text{gen}^\text{ref}$. The objective lowers the denoising loss on $\mathbf{X}^{+}$ relative to $\mathbf{X}^{-}$, measured against the reference model. We add \ac{sft} on the preferred motion and minimize
\begin{equation}\label{eq:post-train}
    \mathcal{L}_{\text{post-train}}=\lambda_{\text{DPO}}\mathcal{L}_\text{DPO}+\lambda_{\text{SFT}}\mathcal{L}_{\text{SFT}}.
\end{equation}
\Cref{sec:post-train-detail} details preference pair construction and post-training.

\subsection{Autoregressive Generation for Video Teleoperation}\label{sec:autoregressive-generation}

For the online setting, we extend \model with an autoregressive motion generator $\pi_\text{gen}^\text{ar}$ that enables segment-by-segment motion generation for real-time video-to-humanoid teleoperation. Each segment has a fixed length of $T$ frames, and consecutive segments share $L$ overlapping frames. Each motion segment $\mathbf{X}^{s}$ is conditioned on the preceding segment $\mathbf{X}^{s-1}$. Following \citet{jiang2024scaling,jiang2024autonomous,jiang2025dynamic}, we inject the overlapping context by replacing the first $L$ frames of the noisy input $\mathbf{X}_n^{s}$ with the last $L$ frames of $\mathbf{X}^{s-1}$. To preserve the ability to generate the initial segment without preceding context, we randomly drop $\mathbf{X}^{s-1}$ during fine-tuning.

During post-training, $\mathbf{X}^{s-1}$ is provided as conditioning only for in-distribution samples with ground-truth motion. For video-only samples, where the preceding ground-truth motion is unavailable, we follow the same post-training procedure as in \cref{sec:post-train} without autoregressive conditioning. Details on the adaptation of the autoregressive generator are in \cref{sec:autoregressive-detail}.

\section{Experiments}

In this section, we first compare \model with reconstruction-retargeting-tracking pipelines and our pre-training-only ablation variants on both datasets with 3D ground-truth annotations and video-only datasets (\cref{sec:comparisons}). We then provide an empirical analysis of DPO post-training, investigating how preference construction and post-training design choices affect humanoid motion executability (\cref{sec:analysis-of-dpo}). We finally study how \model scales with model and data size and whether its post-training gains transfer to other trackers (\cref{sec:scalability}).

\subsection{Comparisons}\label{sec:comparisons}

\paragraph{Datasets.} 
We train $\pi_\text{gen}$ on a mixture of AMASS~\citep{mahmood2019amass}, BONES-SEED~\citep{bonesstudio2026bonesseed}, BEDLAM~\citep{black2023bedlam}, BEDLAM2.0~\citep{tesch2026bedlam2}, and Human3.6M~\citep{ionescu2013human3}. Datasets containing only human motions are retargeted to Unitree G1 motions using our adapted GMR retargeter~\citep{araujo2025retargeting}. During post-training, we fine-tune $\pi_\text{gen}$ on BEDLAM, BEDLAM2.0, Human3.6M, and video-only datasets without motion annotations~\citep{li2020learning,li2025lifting,li2026anylift} covering diverse motion categories, including dancing, fitness, gymnastics, and martial arts. Both tracking policies, $\pi_\text{track}$ and $\pi_\text{track}^\text{priv}$, are fine-tuned exclusively on BONES-SEED.

For evaluation, we use EMDB~\citep{kaufmann2023emdb} and RICH~\citep{huang2022capturing}, whose 3D ground-truth human motions are retargeted to Unitree G1 motions and filtered accordingly. We further conduct evaluation on held-out splits of the video-only datasets. Details on dataset statistics, human-to-humanoid conversion, and conversions across human body models are in \cref{sec:dataset-and-data-conversion-detail}.

\paragraph{Evaluation metrics.} 
For datasets with retargeted 3D ground truth, we report local \ac{mpjpe}, the \ac{pampjpe}, and the \ac{rte}. We also report motion jitter (Jitter), and self-penetration (Pen.) to further evaluate motion quality. For video-only datasets, we report Jitter and Pen., and conduct a \ac{2afc} study comparing the offline generation mode of $\pi_\text{gen}$ against the baselines and ablation variants. Details on the human study are in \cref{sec:human-study-detail}.

We further evaluate the generation-and-tracking pipeline of \model by executing the predicted humanoid motions in MuJoCo~\citep{todorov2012mujoco} with $\pi_\text{track}$. We report the tracking success rate (Succ.) under the termination criterion of \cref{sec:post-train-detail} for all evaluation datasets. For datasets with retargeted ground-truth humanoid motions, we report \ac{mpjpe} and \ac{pampjpe} against the ground truth. Additionally, we report Trk.\ \ac{mpjpe} and Trk.\ \ac{pampjpe}, defined as the \ac{mpjpe} and \ac{pampjpe} between the predicted motions and their corresponding tracked motions, respectively, to measure how easily the motions can be tracked. Details on metrics are in \cref{sec:metric-detail}.

\paragraph{Baselines and ablation variants.} 
We compare against reconstruction-retargeting-tracking pipelines that reconstruct human motion with GVHMR~\citep{shen2024world} and GEM~\citep{li2025genmo}, retarget it with GMR~\citep{araujo2025retargeting}, and track it with $\pi_\text{track}$. For each reconstruction method, we also report a variant retrained on our training data with the same image features and a model size comparable to $\pi_\text{gen}$. Camera motion is estimated with DPVO~\citep{teed2023deep} for GVHMR and \model and with DROID-SLAM~\citep{teed2021droid} for GEM. As ablations, we evaluate the offline generator $\pi_\text{gen}$ and its autoregressive variant $\pi_\text{gen}^\text{ar}$, both without post-training, denoted as ``Ours w/o PT''.

For the executability evaluation in MuJoCo, since $\pi_\text{track}$ follows SONIC~\citep{luo2026sonic} in accepting human motions directly as tracking references, we evaluate the motion reconstruction baselines without retargeting. We additionally report tracking performance on the retargeted ground-truth humanoid motions as a reference. Details on baseline adaptation are in \cref{sec:baseline-detail}.

\paragraph{Results.} 
We report quantitative results on humanoid reference motion quality in \cref{tbl:comparison}. Our offline motion generator consistently outperforms the reconstruction-and-retargeting baselines across motion accuracy and quality metrics, showing the advantage of directly predicting motions in the G1 embodiment. The human study on video-only datasets further shows that participants consistently prefer our results over the baselines, aligning well with the quantitative metrics. Additionally, although the online generator is less competitive on root translation accuracy under the causal setting, it still achieves lower \ac{mpjpe} and \ac{pampjpe} than all pipelines, indicating that accurate local motion can be preserved in the causal setting. Finally, comparisons with the w/o PT variants show that post-training improves motion quality, as reflected by lower motion jitter and self-penetration in most settings, while maintaining comparable motion prediction accuracy.

\Cref{tbl:comparison-track} compares executability. Both post-trained generators achieve higher tracking success than all pipelines on every test set, and the tracked motions of the offline generator are also closer to the ground truth than those of any pipeline. After post-training, $\pi_\text{track}$ follows the references of both generators more closely than the retargeted ground truth, with lower Trk.\ \ac{mpjpe} and Trk.\ \ac{pampjpe} than GT + GMR on EMDB and RICH, and their success rates approach those of GT + GMR (90.2\% and 88.5\% against 91.0\% on EMDB). Post-training raises or preserves the success rate and lowers both tracking errors on all three test sets, with the largest gains for the online generator and on the video-only test set, where the success rate rises from 74.8\% to 80.0\% offline and from 49.6\% to 63.8\% online.

To isolate the effect of camera estimation and assess the upper-bound performance of the approaches, we repeat both comparisons with ground-truth cameras in \cref{sec:comparison-ground-truth-camera}. The resulting trends are consistent with our main experiments. Qualitative comparisons with baselines and ablation variants, along with real-world deployment demonstrations, are available on our \href{https://awfuact.github.io/vit-h/}{\textit{project page}}.

\begin{table*}[t!]
    \centering
    \scriptsize
    \setlength{\tabcolsep}{4.5pt}
    \caption{\textbf{Quantitative comparison of motion quality.} Results use estimated camera motion. $\dagger$~marks baselines retrained on our training data with a model size comparable to \model, and w/o PT denotes a generator before post-training. The \acs{2afc} column reports the percentage of trials in which participants preferred each method over Ours (offline). Lower is better for all metrics except \acs{2afc}. Bold and underline mark the best and second-best results.}
    \label{tbl:comparison}
    \begin{tabular}{lccccccccccccc}
        \toprule
        \multirow{3}{*}{Method} & \multicolumn{5}{c}{EMDB} & \multicolumn{5}{c}{RICH} & \multicolumn{3}{c}{Video-only} \\
        \cmidrule(lr){2-6} \cmidrule(lr){7-11} \cmidrule(lr){12-14}
        & \multirow{2}{*}{\acs{mpjpe}} & PA- & \multirow{2}{*}{\acs{rte}} & \multirow{2}{*}{Jitter} & \multirow{2}{*}{Pen.} & \multirow{2}{*}{\acs{mpjpe}} & PA- & \multirow{2}{*}{\acs{rte}} & \multirow{2}{*}{Jitter} & \multirow{2}{*}{Pen.} & \multirow{2}{*}{Jitter} & \multirow{2}{*}{Pen.} & \multirow{2}{*}{\acs{2afc}} \\
        & & \acs{mpjpe} & & & & & \acs{mpjpe} & & & & & & \\
        \midrule
        GVHMR + GMR & 73.3 & 54.4 & 3.1 & 18.4 & 2.3 & 51.6 & 44.9 & 4.6 & \underline{7.7} & 1.7 & 15.5 & 1.8 & 17.5 \\
        GVHMR$^\dagger$ + GMR & 71.4 & 49.1 & 3.5 & 16.5 & 1.6 & 69.6 & 51.5 & \underline{4.4} & 10.3 & 2.2 & 16.8 & 2.1 & 23.8 \\
        GEM + GMR & 62.3 & 38.4 & 4.6 & 19.6 & 1.3 & 56.5 & 49.4 & 4.7 & 11.9 & 2.2 & 23.9 & 2.2 & 16.3 \\
        GEM$^\dagger$ + GMR & 72.9 & 51.0 & 4.5 & 21.7 & 2.0 & 70.3 & 51.2 & \underline{4.4} & 20.5 & 2.2 & 25.9 & 1.9 & 17.5 \\
        \midrule
        Ours (online) w/o PT & 55.4 & 36.6 & 8.3 & 25.3 & \underline{0.2} & 44.5 & 34.5 & 6.8 & 24.4 & 0.3 & 24.8 & 1.3 & 22.5 \\
        Ours (online) & 55.5 & 36.8 & 6.4 & 18.5 & \textbf{0.1} & 44.3 & 36.1 & 6.3 & 17.4 & \underline{0.2} & 13.6 & 1.2 & 28.8 \\
        \midrule
        Ours (offline) w/o PT & \textbf{52.9} & \textbf{34.0} & \underline{2.6} & \underline{11.2} & \underline{0.2} & \underline{38.7} & \textbf{28.5} & \underline{4.4} & \textbf{7.2} & \underline{0.2} & \underline{10.6} & \underline{1.0} & 41.3 \\
        Ours (offline) & \underline{53.3} & \underline{34.3} & \textbf{2.5} & \textbf{10.6} & \textbf{0.1} & \textbf{38.4} & \underline{29.4} & \textbf{4.3} & \textbf{7.2} & \textbf{0.1} & \textbf{9.5} & \textbf{0.9} & -- \\
        \bottomrule
    \end{tabular}
\end{table*}

\begin{table*}[t!]
    \centering
    \scriptsize
    \setlength{\tabcolsep}{2.5pt}
    \caption{\textbf{Quantitative comparison of motion executability.} All reference motions are executed by $\pi_\text{track}$ in MuJoCo, using estimated camera motion. GT + GMR executes the retargeted ground-truth motions, and rows without GMR track the reconstructed human motions directly. $\dagger$ and w/o PT follow \cref{tbl:comparison}. Higher Succ.\ and lower errors are better. Bold and underline mark the best and second-best results, excluding GT + GMR.}
    \label{tbl:comparison-track}
    \begin{tabular}{lccccccccccccc}
        \toprule
        \multirow{3}{*}{Method} & \multicolumn{5}{c}{EMDB} & \multicolumn{5}{c}{RICH} & \multicolumn{3}{c}{Video-only} \\
        \cmidrule(lr){2-6} \cmidrule(lr){7-11} \cmidrule(lr){12-14}
        & \multirow{2}{*}{Succ.} & \multirow{2}{*}{\acs{mpjpe}} & PA- & Trk. & Trk.\ PA- & \multirow{2}{*}{Succ.} & \multirow{2}{*}{\acs{mpjpe}} & PA- & Trk. & Trk.\ PA- & \multirow{2}{*}{Succ.} & Trk. & Trk.\ PA- \\
        & & & \acs{mpjpe} & \acs{mpjpe} & \acs{mpjpe} & & & \acs{mpjpe} & \acs{mpjpe} & \acs{mpjpe} & & \acs{mpjpe} & \acs{mpjpe} \\
        \midrule
        GT + GMR & 91.0 & 35.2 & 22.7 & 35.2 & 22.7 & 100.0 & 30.3 & 19.6 & 30.3 & 19.6 & -- & -- & -- \\
        \midrule
        GVHMR + GMR & 61.7 & 101.2 & 59.8 & 47.5 & 29.8 & 90.3 & \underline{64.6} & 47.2 & 36.4 & 28.9 & 50.3 & 91.9 & 52.2 \\
        GVHMR$^\dagger$ + GMR & 71.4 & 94.5 & 57.0 & 44.7 & 30.0 & 84.7 & 90.8 & 58.2 & 53.4 & 34.7 & 48.7 & 105.8 & 60.3 \\
        GEM + GMR & 73.0 & 94.9 & 43.0 & 43.2 & 25.4 & 90.3 & 71.7 & 51.0 & 37.9 & 29.7 & 51.7 & 99.8 & 57.2 \\
        GEM$^\dagger$ + GMR & 73.0 & 100.0 & 58.1 & 43.9 & 29.4 & 90.3 & 90.9 & 55.1 & 52.5 & 34.2 & 46.7 & 106.8 & 59.8 \\
        GVHMR & 56.4 & 91.3 & 45.7 & -- & -- & 29.2 & 83.5 & 46.4 & -- & -- & 20.5 & -- & -- \\
        GVHMR$^\dagger$ & 54.9 & 85.8 & 45.7 & -- & -- & 26.4 & 82.0 & 46.5 & -- & -- & 21.9 & -- & -- \\
        GEM & 54.9 & 94.3 & 45.8 & -- & -- & 30.6 & 86.8 & 47.5 & -- & -- & 17.3 & -- & -- \\
        GEM$^\dagger$ & 55.6 & 90.1 & 45.8 & -- & -- & 23.6 & 83.7 & 45.9 & -- & -- & 19.5 & -- & -- \\
        \midrule
        Ours (online) w/o PT & 76.9 & 85.8 & 41.3 & 34.0 & 23.1 & 87.5 & 76.0 & 40.9 & 34.3 & 21.6 & 49.6 & 71.7 & 41.9 \\
        Ours (online) & 88.5 & 84.0 & 40.7 & \textbf{29.4} & \textbf{16.4} & \underline{98.2} & 69.8 & 40.0 & 27.6 & \textbf{17.3} & 63.8 & 65.2 & 37.8 \\
        \midrule
        Ours (offline) w/o PT & \underline{89.5} & \underline{75.4} & \underline{40.0} & 32.0 & 18.6 & \textbf{100.0} & \textbf{62.0} & \textbf{34.3} & \underline{26.3} & 18.2 & \underline{74.8} & \underline{59.0} & \underline{37.1} \\
        Ours (offline) & \textbf{90.2} & \textbf{75.1} & \textbf{39.7} & \underline{31.5} & \underline{18.0} & \textbf{100.0} & \textbf{62.0} & \underline{34.6} & \textbf{25.9} & \underline{17.7} & \textbf{80.0} & \textbf{54.2} & \textbf{34.7} \\
        \bottomrule
    \end{tabular}
\end{table*}

\begin{figure*}[t!]
    \centering
    \includegraphics[width=\linewidth]{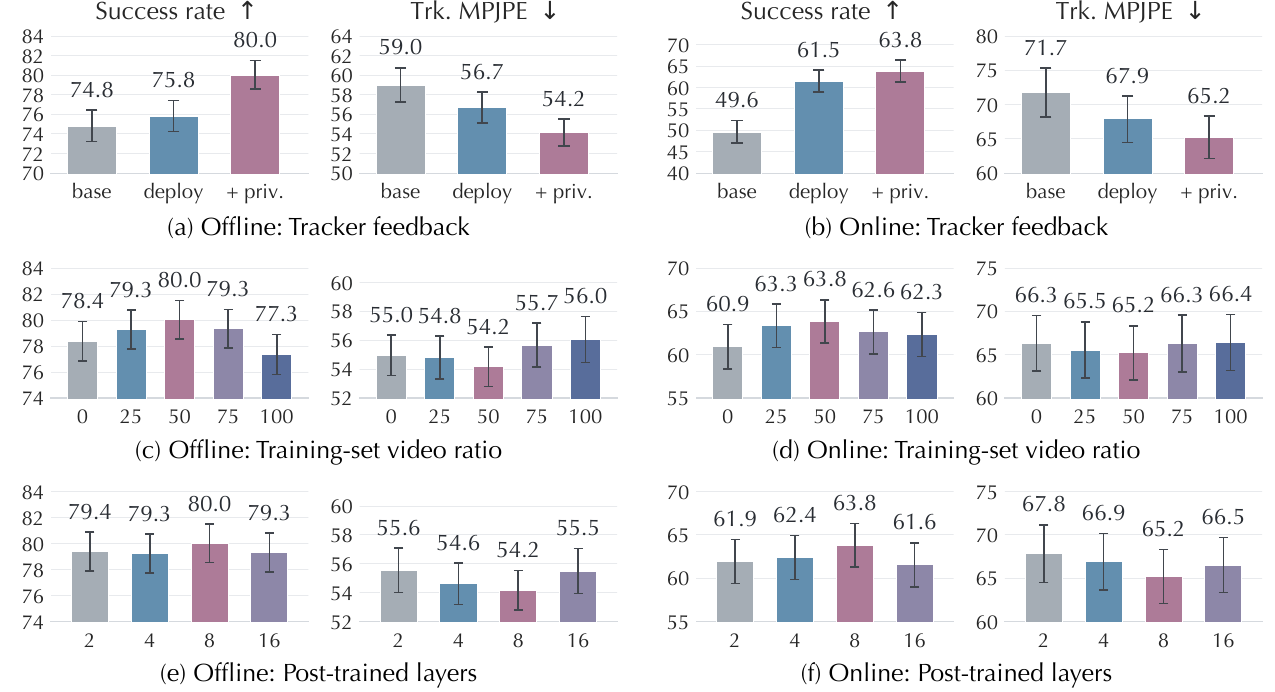}
    \caption{\textbf{Analysis of \acs{dpo} post-training on the video-only test set.} Panels (a,b) vary the tracker feedback, from the pre-trained generator (base) to post-training with rewards from the deployable tracker alone (deploy) and with the privileged tracker added (+ priv.). Panels (c,d) vary the amount of in-distribution pre-training set data relative to video-only data during post-training, and panels (e,f) the number of final Transformer layers updated during post-training. Error bars denote 95\% confidence intervals.}
    \label{fig:dpo_ablation}
\end{figure*}

\begin{figure*}[t!]
    \centering
    \includegraphics[width=\linewidth]{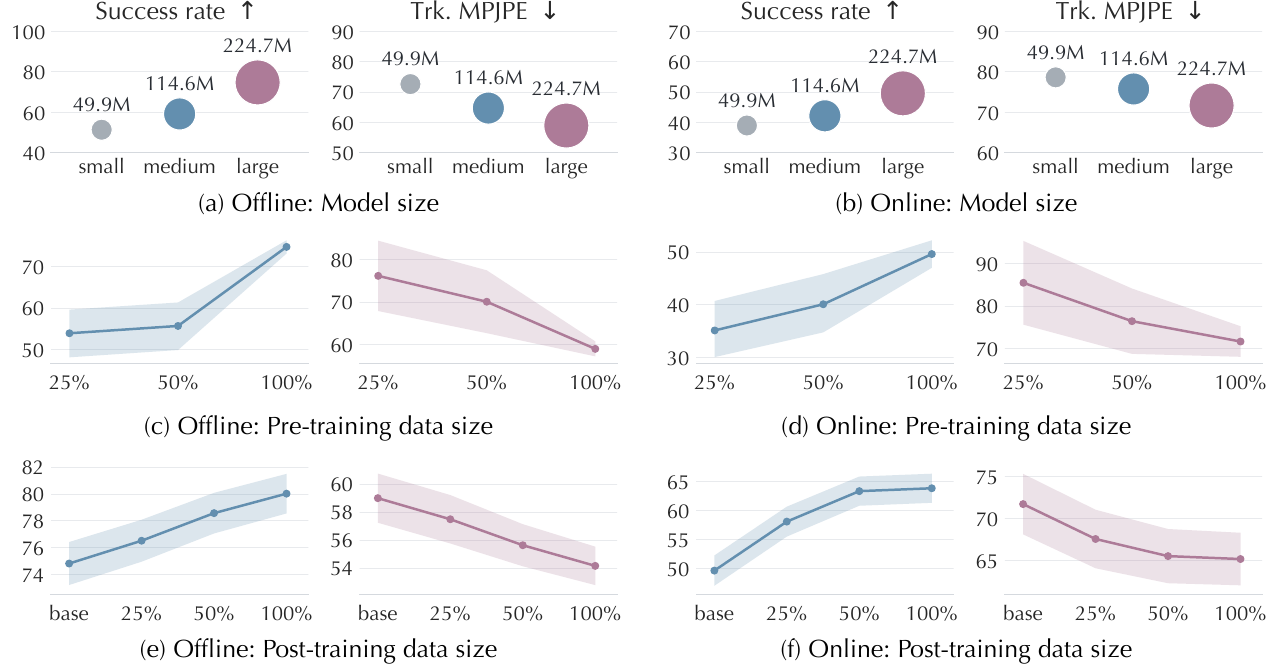}
    \caption{\textbf{Scaling behavior of \model on the video-only test set.} Panels (a,b) vary the model size, with marker size indicating the number of parameters, panels (c,d) the percentage of pre-training data, and panels (e,f) the percentage of post-training data, where base denotes the pre-trained generator. Panels (a) to (d) use generators without post-training. Shaded regions denote 95\% confidence intervals.}
    \label{fig:scalability}
\end{figure*}

\subsection{Analysis of \texorpdfstring{\ac{dpo}}{DPO} Post-Training}\label{sec:analysis-of-dpo}

We analyze three aspects of post-training: reward design, post-training data sources, and fine-tuning depth. Since EMDB and RICH are less challenging under our evaluation protocol (\cref{tbl:comparison-track}), we focus this analysis on the more challenging video-only test set. Results are shown in \cref{fig:dpo_ablation}.

First, rewards from the deployable tracker alone raise the success rate from 74.8\% to 75.8\% for the offline generator and from 49.6\% to 61.5\% for the online generator, and adding the global-motion reward from the privileged tracker $\pi_\text{track}^\text{priv}$ raises it further to 80.0\% and 63.8\%. Since all evaluations use the deployable tracker, this gain indicates that global tracking feedback supplies supervision that local feedback misses. Second, we vary the amount of in-distribution pre-training set data relative to the video-only out-of-distribution data. Performance exhibits a non-monotonic trend, with the best empirical results achieved when the amount of in-distribution data is half that of video-only data, while further increasing the amount of pre-training set data degrades performance. Finally, we vary the number of Transformer layers updated during DPO post-training. Among the 16 layers of our generator, fine-tuning the last 8 yields the best empirical performance, while updating either fewer layers or all 16 layers results in lower performance.

\subsection{Scalability and Cross-Tracker Generalization}\label{sec:scalability}

\paragraph{Scaling with model and data size.}
We study how \model scales with model capacity and training data on the video-only test set. Results are reported in \cref{fig:scalability}. Model-size and pre-training-data results are obtained without post-training. Increasing the generator from 49.9M to 114.6M and 224.7M parameters consistently raises the tracking success rate and lowers Trk.\ \ac{mpjpe} for both the offline and the online generator. Pre-training on 25\%, 50\%, and 100\% of the pre-training data shows the same trend, with the largest gain in success rate from 50\% to 100\%. Post-training data scale behaves similarly. Increasing post-training data consistently improves both metrics for both the offline and the online generators.

\begin{figure*}[t!]
    \centering
    \includegraphics[width=\linewidth]{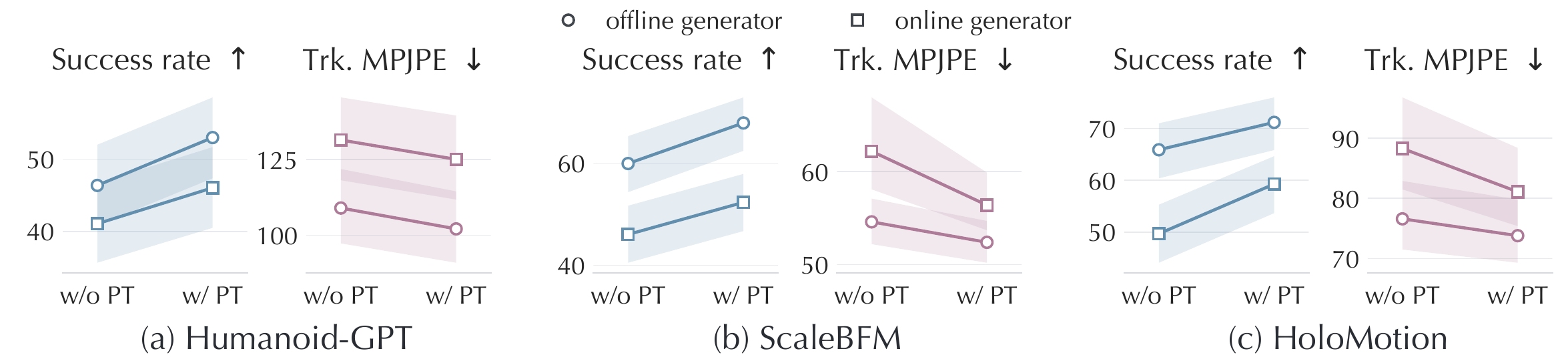}
    \caption{\textbf{Post-training gains transfer to trackers not used in post-training.} Motions generated by the offline (circles) and autoregressive (squares) generators on the video-only test set are executed by (a) Humanoid-GPT, (b) ScaleBFM, and (c) HoloMotion, before (w/o PT) and after (w/ PT) post-training. Shaded regions denote 95\% confidence intervals.}
    \label{fig:tracker}
\end{figure*}

\paragraph{Generalization across trackers.}
To test whether post-training merely fits $\pi_\text{track}$, we execute the motions of both generators on the video-only test set with three trackers not used in post-training, Humanoid-GPT~\citep{qi2026humanoid}, ScaleBFM~\citep{zeng2026scaling}, and HoloMotion~\citep{chen2026holomotion}. As \cref{fig:tracker} shows, post-training raises the success rate and lowers Trk.\ \ac{mpjpe} for both generators with all three trackers, indicating that it improves the executability of the generated motions rather than overfitting to the tracker that provides the feedback.

\section{Conclusion}

We presented \model, which generates humanoid reference motions from monocular video directly in the motion space of the Unitree G1 embodiment and post-trains the generator on the outcomes of the tracker that executes them. This design yields reference motions that are more executable, and more accurate in local pose, than those of state-of-the-art reconstruction-retargeting-tracking pipelines. Post-training on tracking feedback, which does not necessarily require motion annotations, brings its largest gains on held-out internet videos, and these gains carry over to trackers never used for feedback. \model also scales with model and data size and, through its autoregressive variant, supports real-time teleoperation of a physical humanoid. Together, these results suggest that as general-purpose trackers mature, the tracker can serve not only as the executor of reference motions but also as the supervisor of the model that produces them. 

\paragraph{Limitations and future work.} Our scaling study is limited by available training data and computational resources, while our current framework does not model dexterous hand poses, restricting fine-grained object manipulation. Future work includes scaling to larger models and datasets, exploring alternative post-training algorithms and feedback from multiple trackers, and extending the framework to support dexterous manipulation. A comprehensive discussion on limitations and future directions is provided in \cref{sec:limitations-and-future-work}.

\subsection*{Acknowledgments}

We thank Long Liu, Yunhao Li, Yuyang Li, and Zhenghao Qi for hardware support. H. Li thanks Jiaman Li, Yanzhe Lyu, and \v{Z}iga Kova\v{c}i\v{c} for fruitful discussions. This work is supported in part by the Brain Science and Brain-like Intelligence Technology -- National Science and Technology Major Project (2025ZD0219400), the National Natural Science Foundation of China (62376009), the Beijing Nova program, the NVIDIA Academic Grant Program using Spark and Thor, the PKU-BingJi Joint Laboratory for Artificial Intelligence, the Wuhan Major Scientific and Technological Special Program (2025060902020304), the Hubei Embodied Intelligence Foundation Model Research and Development Program, and the National Comprehensive Experimental Base for Governance of Intelligent Society, Wuhan East Lake High-Tech Development Zone.

\subsection*{AI use statement}

In this work, we used generative AI tools to assist with implementing methods under the authors' supervision. We have not used generative AI tools to help develop theoretical models or conceptual frameworks, propose or refine hypotheses, design or provide feedback on research methodology or experiments, assist with translation, clean or reformat datasets, support qualitative or thematic data analysis, or interpret results. Our work introduces no synthetic datasets and contains no mathematical claims or proofs, so the corresponding disclosures do not apply.

Additionally, we used generative AI tools to edit the paper for readability, search for information, and identify relevant literature. We have reviewed all AI-assisted work. Specifically, the co-first authors reviewed, tested, and verified the correctness of all AI-assisted method implementations. We also carefully reviewed all AI-assisted edits to the manuscript and manually verified the relevance and correctness of the identified literature.

\subsection*{Ethics statement}

The human study (\cref{sec:human-study-detail}) involved 20 adult volunteers who gave informed consent before participating, and no personally identifiable information was collected. The teleoperators shown in our figures and videos consented to being recorded, and their faces are obscured in the figures. The internet videos used for post-training and evaluation come from existing research datasets~\citep{li2020learning,li2025lifting,li2026anylift}.

\subsection*{Reproducibility statement}

The appendix provides the details needed to reproduce our method and experiments. \Cref{sec:ours-details} describes the data representation, network architecture, pre-training, and post-training of the motion generator, the adaptation of the autoregressive generator, the training of the motion trackers, and the real-world deployment setup. \Cref{sec:experiments-details} describes the dataset statistics and data conversion, the human study, the evaluation metrics, and the adaptation and training of the baselines. We will release our code and processed datasets.

\bibliography{references}
\bibliographystyle{iclr2027_conference}

\clearpage
\acresetall
\appendix
\crefalias{section}{appendix}
\crefalias{subsection}{appendix}
\renewcommand\thefigure{A\arabic{figure}}
\setcounter{figure}{0}
\renewcommand\thetable{A\arabic{table}}
\setcounter{table}{0}
\renewcommand\theequation{A\arabic{equation}}
\setcounter{equation}{0}
\pagenumbering{arabic}
\renewcommand*{\thepage}{A\arabic{page}}
\setcounter{footnote}{0}
\addtocontents{toc}{\protect\setcounter{tocdepth}{2}}

\hypersetup{linkcolor=black}
\renewcommand{\contentsname}{Supplementary Contents}
\tableofcontents
\hypersetup{linkcolor=red}

\section{Overview}

This supplementary material provides further details and additional results for the experiments presented in the main paper. We first describe additional implementation details of \model (\cref{sec:ours-details}) and the experimental setup (\cref{sec:experiments-details}). We then provide additional experimental results (\cref{sec:experiments-results}). We conclude with limitations and directions for future work (\cref{sec:limitations-and-future-work}). We highly recommend viewing our \href{https://awfuact.github.io/vit-h/}{\textit{project page}} for demonstrations across diverse scenarios.

\section{Details on \modelname}\label{sec:ours-details}

In this section, we provide additional details on the data representation (\cref{sec:data-representation-detail}), network architecture (\cref{sec:network-architecture-detail}), pre-training and post-training of our reference motion generator (\cref{sec:motion-generator-detail,sec:post-train-detail}). We further describe the adaptations to the reference motion generator for its autoregressive variant (\cref{sec:autoregressive-detail}). Finally, we provide details on the training of our motion trackers (\cref{sec:motion-tracker-detail}) and the deployment of our framework in both offline and online settings (\cref{sec:deploy-detail}).

\subsection{Data Representation of Motion Generator}\label{sec:data-representation-detail}

\paragraph{Video-derived conditions.} 
We represent each video using relative camera orientations, 2D bounding boxes, 2D keypoints, and image features. We estimate relative camera orientations with DPVO~\citep{teed2023deep}, detect human bounding boxes with YOLOv8~\citep{jocher2023ultralytics}, extract human keypoints with ViTPose~\citep{xu2022vitpose}, and extract image features with Fast SAM 3D Body~\citep{yang2026fastsam3dbody}, an accelerated variant of SAM 3D Body~\citep{yang2026sam3dbody}. For datasets without paired videos, we simulate camera trajectories, generate 2D bounding boxes, project 3D human motions into the image plane to obtain 2D keypoints, and set the image features to zero.

\paragraph{Global humanoid motion recovery.} After predicting the humanoid motion represented by $\mathbf{X}=(\mathbf{v}_\text{root},\boldsymbol{\Gamma}_\text{cam},\boldsymbol{\Gamma}_\text{gv},\mathbf{Q},\boldsymbol{\tau},\mathbf{p}_\text{ee})$ as described in \cref{sec:motion-generation}, we discard $\boldsymbol{\tau}$ and $\mathbf{p}_\text{ee}$, which are used only for loss computation, and recover the global root transformation following prior human motion reconstruction works that adopt the gravity-view coordinate system~\citep{shen2024world,li2025genmo}. We then obtain the joint and vertex positions in world coordinates through forward kinematics using the Unitree G1 kinematic model.

\subsection{Network Architecture of Motion Generator}\label{sec:network-architecture-detail}

We adopt a Transformer backbone~\citep{vaswani2017attention} with rotary positional embeddings (RoPE)~\citep{su2024roformer} for the video-conditioned reference motion generator $\pi_\text{gen}$. The conditions are independently projected by MLP encoders and then concatenated with the noisy motion before being fed into the backbone. The Transformer outputs are then projected by an output MLP to obtain the denoised motion sequence. 

Specifically, the backbone of the reference motion generator consists of 16 Transformer encoder layers with 16 attention heads and a hidden dimension of 1,024, using GELU activations~\citep{hendrycks2016gaussian}, resulting in 224.7M parameters in total. For the model-scaling experiments conducted in \cref{sec:scalability}, we additionally use a small variant with 12 layers, 8 attention heads, and a hidden dimension of 512 (49.9M parameters), and a medium variant with 14 layers, 12 attention heads, and a hidden dimension of 768 (114.6M parameters).

\subsection{Pre-Training of Motion Generator}\label{sec:motion-generator-detail}

\paragraph{Training losses.}
Beyond the diffusion loss described in \cref{eq:diffusion-loss}, we introduce the following auxiliary losses to facilitate training: (i) A local joint loss, which applies the Huber loss to root-relative joint positions; (ii) A local vertex loss, which applies the Huber loss to root-relative vertex positions; (iii) a global translation loss, which applies the $\ell_\text{1}$ loss to the recovered root translation; and (iv) an end-effector stationary loss, which applies binary cross-entropy loss to the stationary states of the hands and feet. For the local vertex loss, we preselect 384 vertices on the Unitree G1 surface using farthest point sampling, with denser sampling around the hands and feet. For the end-effector stationary loss, the ground-truth stationary labels are determined from the velocities of the corresponding joints.

We additionally include the following 2D reprojection losses during training: (i) a 2D joint loss, which applies the Huber loss to reprojected 2D joint positions; (ii) a 2D vertex loss, which applies the Huber loss to reprojected 2D vertex positions; and (iii) a 2D translation loss, which applies the Huber loss to the root translation in camera coordinates following the formulation of CLIFF~\citep{li2022cliff}. For all 2D losses, we use the cameras associated with the paired 3D human motions. We rescale the camera translation by a factor of $0.8$ to account for the difference in body scale between humans and the Unitree G1 humanoid.

\paragraph{Training setup.} 
Our video-conditioned reference motion generator is trained using a DDPM scheduler~\citep{ho2020denoising} with 1,000 diffusion steps. We use the Adam optimizer~\citep{diederik2015adam} with a fixed learning rate of $2\times10^{-4}$. For both offline and online generation, we train the model with a fixed sequence length of 120 frames at 30 fps. We train all models on 4 NVIDIA H100 GPUs with a batch size of 64 per GPU, yielding a global batch size of 256. All models are pre-trained for 730k optimization steps.

\subsection{Post-Training of Motion Generator}\label{sec:post-train-detail}

\paragraph{Preference pair construction.} 
For each video-derived condition set $\mathcal{C}$, we construct a candidate pool by sampling $M=128$ reference motions, $\{\mathbf{X}^{(i)}\}_{i=1}^{M}\sim\pi_{\text{gen}}(\cdot\mid\mathcal{C})$, using DDIM sampling~\citep{song2021denoising} with 50 denoising steps.

For a humanoid motion sequence $\mathbf{X}=(\mathbf{v}_\text{root},\boldsymbol{\Gamma}_\text{cam},\boldsymbol{\Gamma}_\text{gv},\mathbf{Q},\boldsymbol{\tau},\mathbf{p}_\text{ee})$ generated by our reference motion generator $\pi_\text{gen}$, we denote the recovered root translation, root orientation, and joint positions (including the root) in world coordinates as $\mathbf{t}\in\mathbb{R}^{T\times3}$, $\mathbf{R}\in\mathbb{R}^{T\times3\times3}$, and $\mathbf{J}\in\mathbb{R}^{T\times (J+1)\times3}$, respectively (\cref{sec:data-representation-detail}). We further convert the output 6D local joint rotation $\mathbf{Q}$ into 1D joint angle $\mathbf{q}\in\mathbb{R}^{T\times J\times1}$. For two humanoid motions $\mathbf{X}$ and $\mathbf{X}'$, we compute:
\begin{align}
    \mathcal{E}_{\text{root-pos}}(\mathbf{X},\mathbf{X}')&=\frac{1}{T}\sum_{t}\big\Vert(\mathbf{t}_t-\mathbf{t}_0)-(\mathbf{t}'_t-\mathbf{t}'_0)\big\Vert_2^2, \\
    \mathcal{E}_{\text{root-ori}}(\mathbf{X},\mathbf{X}')&=\frac{1}{T}\sum_{t}\text{d}_{\text{SO(3)}}^2(\mathbf{R}_t,\mathbf{R}'_t),\\
    \mathcal{E}_{\text{pos}}(\mathbf{X},\mathbf{X}')&=\frac{1}{T(J+1)}\sum_{t}\sum_{j}\big\Vert\mathbf{R}_t^\top(\mathbf{J}_{t,j}-\mathbf{t}_{t})-{\mathbf{R}'_t}^\top(\mathbf{J}'_{t,j}-\mathbf{t}'_{t})\big\Vert_2^2,\\
    \mathcal{E}_{\text{ori}}(\mathbf{X},\mathbf{X}')&=\frac{1}{TJ}\sum_{t}\sum_{j}(\mathbf{q}_{t,j}-\mathbf{q}'_{t,j})^2,
\end{align}
where $\text{d}_{\text{SO(3)}}$ denotes the geodesic distance on $\text{SO(3)}$. We then convert each error into a normalized reward using an exponential kernel. The scale factor in \cref{eq:normalized-error} is calibrated from the empirical error distribution as $s_k=\sqrt{\operatorname{Median}(\mathcal{E}_k)}$. We calibrate the scales separately for different data and generation settings to account for differences in their error distributions. 

We assign equal weights to all components of the local pose reward $\mathcal{R}_{\text{local}}$ and global motion reward $\mathcal{R}_{\text{global}}$ defined in \cref{eq:reward-local,eq:reward-global}. For the tracking-success reward $\mathcal{R}_{\text{succ}}$, we determine tracking success following the evaluation termination criterion of SONIC~\citep{luo2026sonic}. Specifically, a rollout is considered failed at frame $t$ if any of the following conditions is met: \begin{equation} 
    \vert z_{\text{root},t}^{\text{ref}}-z_{\text{root},t}^{\text{track}}\vert>0.25\,\text{m},\quad \text{d}_{\text{SO(3)}}(\mathbf{R}_{t}^{\text{ref}},\mathbf{R}_{t}^{\text{track}})>1.0\,\text{rad},\quad \max_{e\in\mathcal{B}_{\text{ee}}}\vert z_{e,t}^{\text{ref}}-z_{e,t}^{\text{track}}\vert>0.25\,\text{m},
\end{equation} 
where $\mathcal{B}_{\text{ee}}$ contains the left and right ankle-roll links and the left and right wrist-yaw links. The rollout is considered successful only if no failure is triggered throughout the motion. We apply the same criterion to obtain $s$ and $s_{\text{priv}}$ for $\pi_{\text{track}}$ and $\pi_{\text{track}}^{\text{priv}}$, respectively (\cref{eq:reward-succ}).

The reward dimensions used for preference-pair construction depend on whether ground-truth humanoid motion is available. In our post-training data, ground-truth humanoid motions are available for BEDLAM~\citep{black2023bedlam}, BEDLAM2.0~\citep{tesch2026bedlam2}, and Human3.6M~\citep{ionescu2013human3}, while the video-only datasets~\citep{li2020learning,li2025lifting,li2026anylift} do not provide ground-truth motions. For videos without ground truth, we treat the root and local components of $r_2^{(i)}$ in \cref{eq:reward-wo-gt} as separate dimensions during preference-pair filtering.

We impose a margin $\boldsymbol{\delta}$ on the reward advantage when constructing preference pairs. We use different margin settings for the two sources of post-training data. For both data sources, the tracking-success reward uses a zero margin with a non-strict constraint, $r_3^{(i)}-r_3^{(j)}\geq0$. For the video-only datasets, we set all continuous reward margins to zero to retain as much internet video data as possible. For videos from the pre-training set, which are more abundant, we use the margins to control the amount of preference data. Specifically, we first identify, for each condition, the candidate pair with the largest weighted reward gap under zero continuous margins. We then compute the reward gaps of these selected pairs along each reward dimension and set the corresponding margin $\delta_k$ to the $p$-th percentile of its gap distribution. We tune the shared percentile $p$ such that the number of retained pre-training conditions is roughly comparable to that of the video-only datasets.

We finally select the preference pair from all candidate pairs satisfying the margin constraints by maximizing the weighted reward gap, as described in \cref{eq:pair-select}. We assign a weight of $10$ to the tracking-success reward and a weight of $1$ to all other reward dimensions.

\paragraph{Training setup.}
For post-training, we set $\beta_{\text{DPO}}=10$, $\lambda_{\text{DPO}}=0.05$, and $\lambda_{\text{SFT}}=2$ for the objective defined in \cref{eq:dpo,eq:post-train}. The SFT objective is applied to the preferred motion $\mathbf{X}^{+}$. We use the AdamW optimizer~\citep{loshchilov2019decoupled} with a fixed learning rate of $1\times10^{-4}$, weight decay of $0.01$, and gradient norm clipping at $1.0$. Unlike pre-training, we do not constrain the motion sequences to a fixed length during post-training. We post-train all models on 4 NVIDIA H100 GPUs with a batch size of 16 per GPU, yielding a global batch size of 64. All models are post-trained for 30k optimization steps.

\subsection{Adaptation of Autoregressive Motion Generator}\label{sec:autoregressive-detail}

\paragraph{Pre-training setup.} 
For the autoregressive generator $\pi_\text{gen}^\text{ar}$, we use the same sequence length of $T=120$ frames and sample the overlap length from $L\in\{0, T-5\}$. When $L=T-5$, the first $T-5$ frames of the noisy motion input are replaced by the corresponding frames from the preceding segment, while $L=0$ corresponds to generation without preceding-motion conditioning. We use two learnable conditioning tokens to distinguish these two cases, selecting the corresponding token based on whether preceding-motion conditioning is provided. The remaining diffusion scheduler, optimizer, and training configurations follow those of the offline generator (\cref{sec:motion-generator-detail}).

\paragraph{Post-training setup.} For post-training $\pi_\text{gen}^\text{ar}$, we follow the same DPO objective and training setup (\cref{sec:post-train-detail}). To match the deployment setting (\cref{sec:deploy-detail}), we sample candidate motions using DDIM with 4 denoising steps during post-training. For in-distribution videos from the pre-training datasets with ground-truth motion, we include both samples with and without preceding-motion conditioning. The former use fixed-length sequences of $T=120$ frames with ground-truth preceding motion, while the latter retain their original sequence lengths without autoregressive conditioning. Internet video samples are similarly used without preceding-motion conditioning and retain their original lengths. We keep the numbers of the two types of pre-training-dataset samples balanced, and their combined duration roughly comparable to that of the Internet video samples.

\paragraph{Autoregressive inference.} During online generation, each segment is conditioned on the preceding $T-5$ frames and generates 5 new frames. We maintain a global humanoid motion queue containing all previously generated motions. Before generating $\mathbf{X}^{s}$, we canonicalize the latest $T-5$ frames in the queue to construct the preceding motion context $\mathbf{X}^{s-1}$. After generation, we align the 5 newly generated frames to the global motion queue using the relative root transformation between the last context frame, i.e., the sixth frame from the end of $\mathbf{X}^{s}$, and its corresponding frame in the global queue, and then append the aligned frames to the queue. For long-sequence video-to-humanoid teleoperation,  we apply a lightweight damping operation to the root transformation of each generated segment to reduce accumulated drift.

\subsection{Training of Motion Tracker}\label{sec:motion-tracker-detail}

\paragraph{Training of the deployable motion tracker.}
We fine-tune SONIC~\citep{luo2026sonic} into $\pi_\text{track}$ with \acl{ppo} \citep[PPO;][]{schulman2017proximal}\acused{ppo}. The added rewards track the horizontal position and the yaw of the anchor body, each with a weight of 0.5 and a Gaussian kernel with a standard deviation of 0.5\,m and 0.6\,rad, respectively. For \ac{ppo}, we set the actor learning rate to $5\times10^{-6}$ with an adaptive range of $[5\times10^{-6},1\times10^{-5}]$ and reduce the desired KL divergence from 0.01 to 0.005.

We add 108 dimensions of privileged global trajectory information to the critic observations, namely the current root transformations of the robot and the reference motion, together with the reference root transformations of the next 10 frames, which increases the critic observation dimension from 1,645 to 1,753 and leaves the actor observation unchanged. To initialize the expanded critic from the released SONIC checkpoint, we copy the weights for the original observations and zero-initialize those for the new observations, whose normalization statistics start from zero mean and unit variance. We train $\pi_\text{track}$ for 20k steps on 8 NVIDIA H100 GPUs.

\paragraph{Training of the privileged motion tracker.}
We initialize $\pi_\text{track}^\text{priv}$ from $\pi_\text{track}$ and append the 3D relative root position and 6D relative root orientation between the robot and the reference motion to the actor observations, which increases the actor observation dimension from 930 to 939. The expanded actor copies the weights for the original observations and zero-initializes those for the new ones. We freeze the SONIC encoders and the auxiliary kinematic decoder and optimize only the action decoder and the critic, keeping the rewards, critic observations, and \ac{ppo} hyperparameters of $\pi_\text{track}$. We train $\pi_\text{track}^\text{priv}$ for 200k steps on 8 NVIDIA H100 GPUs.

\subsection{Real-World Deployment}\label{sec:deploy-detail}

\paragraph{Deployment setup with offline motion generator.}
Consistent with our post-training setup (\cref{sec:post-train-detail}), we use DDIM sampling with 50 denoising steps for the offline reference motion generator $\pi_\text{gen}$. We apply no post-processing to the generated motions and directly track them with the deployable motion tracker $\pi_\text{track}$ for real-world deployment.

\begin{figure*}[t!]
    \centering
    \includegraphics[width=\linewidth]{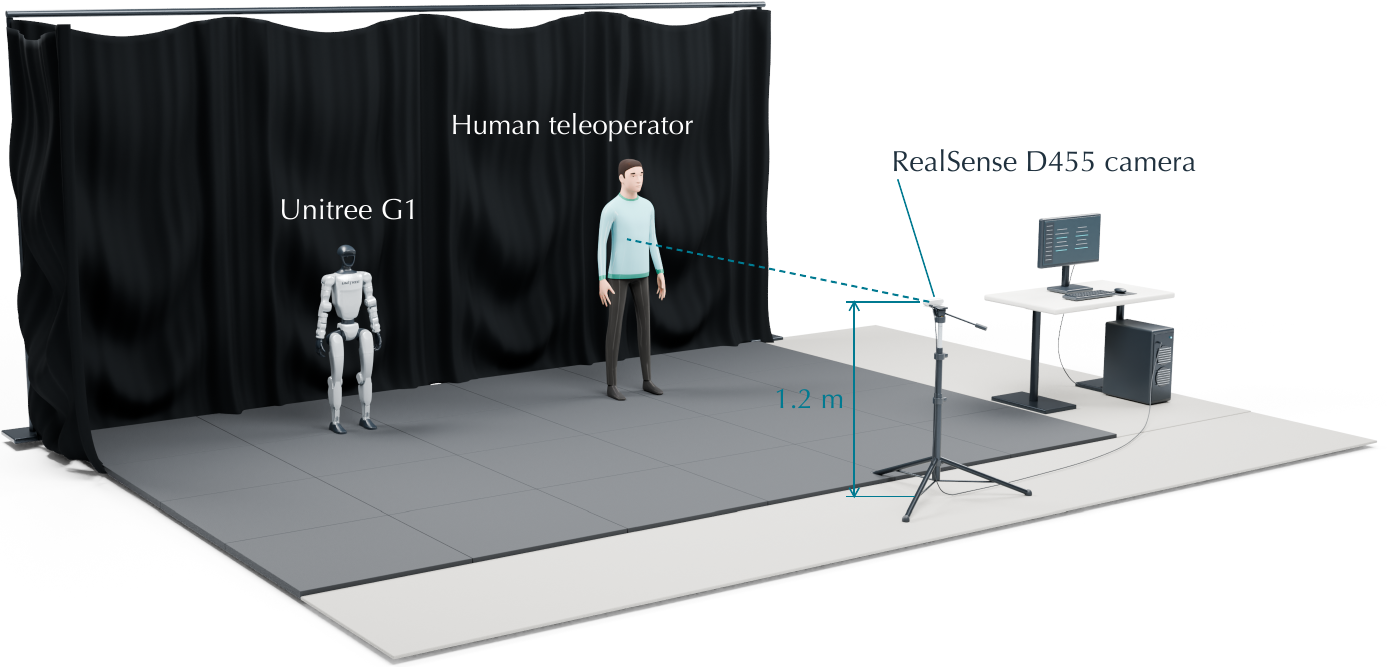}
    \caption{\textbf{Video teleoperation setup.} A RealSense D455 camera mounted 1.2\,m above the ground, with its optical axis approximately parallel to the ground, records RGB video of the teleoperator, and the Unitree G1 stands beside the teleoperator.}
    \label{fig:teleoperation}
\end{figure*}

\begin{table}[t!]
    \centering
    \small
    \caption{\textbf{Latency breakdown of real-time teleoperation.} We report the average and peak latency of each module in ms, measured on the NVIDIA RTX 5090 GPU. The total is the sum of the average latencies. Peak latencies are measured independently for each module and are therefore not summed.}
    \label{tbl:computational}
    \setlength{\tabcolsep}{13pt}
    \resizebox{\linewidth}{!}{
        \begin{tabular}{lccccc}
            \toprule
            \multirow{2}{*}{Latency (ms)} & Frame & Video & Motion & Motion & \multirow{2}{*}{Total} \\
            & transmission & preprocessing & generation & tracking & \\
            \midrule
            Average & 70.1 & 114.4 & 62.3 & 6.9 & 253.7 \\
            Peak & 130.8 & 232.1 & 78.2 & 12.0 & -- \\
            \bottomrule
        \end{tabular}
    }
\end{table}

\paragraph{Deployment setup with online motion generator.} For real-time video-to-humanoid teleoperation in the online setting, we use DDIM sampling with 4 denoising steps for our autoregressive reference motion generator $\pi_\text{gen}^\text{ar}$ to reduce inference latency.

We next describe the video teleoperation setup, illustrated in \cref{fig:teleoperation}. We use an Intel RealSense D455 camera and only use its RGB stream as input, without depth. The camera captures 640$\times$480 images at 30 Hz and is mounted 1.2 m above the ground, with its viewing direction parallel to the ground, to capture the human teleoperator. Each incoming frame is processed online to extract the video-derived conditions (\cref{sec:data-representation-detail}). Since the camera remains fixed throughout teleoperation, we do not estimate relative camera orientations. 

At system initialization, approximately 10 seconds are required to generate the first motion segment without preceding motion context. Afterwards, the system operates continuously by conditioning on the preceding $T-5$ frames and generating 5 new frames at a time. Once the preprocessing results for 5 new video frames are available, they are passed to $\pi_\text{gen}^\text{ar}$ to generate the corresponding reference motion, which is appended to a reference-motion buffer for tracking. The buffered motion is then interpolated to 50 fps before being passed to the deployable tracker $\pi_\text{track}$. Since $\pi_\text{track}$ inherits SONIC's 10-frame future reference input, we repeat the current reference frame to avoid additional latency. We provide a computational timing breakdown in \cref{tbl:computational}. Our system achieves an average end-to-end latency below 300 ms. We further reserve a 100 ms latency margin, enabling stable real-time video-to-humanoid teleoperation with sub-400 ms latency.

\section{Details on Experiments}\label{sec:experiments-details}

In this section, we provide additional details on our experimental setup. Throughout our experiments, we use the 29-DoF Unitree G1 as the common humanoid platform for both simulation and real-world deployment. We first present statistics of the pre-training and post-training datasets and describe the data conversion procedures across different representations (\cref{sec:dataset-and-data-conversion-detail}). We then provide details on the human study setup (\cref{sec:human-study-detail}) and the formulation of our evaluation metrics (\cref{sec:metric-detail}). We finally detail our adaptations of the baseline methods (\cref{sec:baseline-detail}).

\subsection{Dataset Statistics and Data Conversion}\label{sec:dataset-and-data-conversion-detail}

\begin{figure*}[t!]
    \centering
    \includegraphics[width=\linewidth]{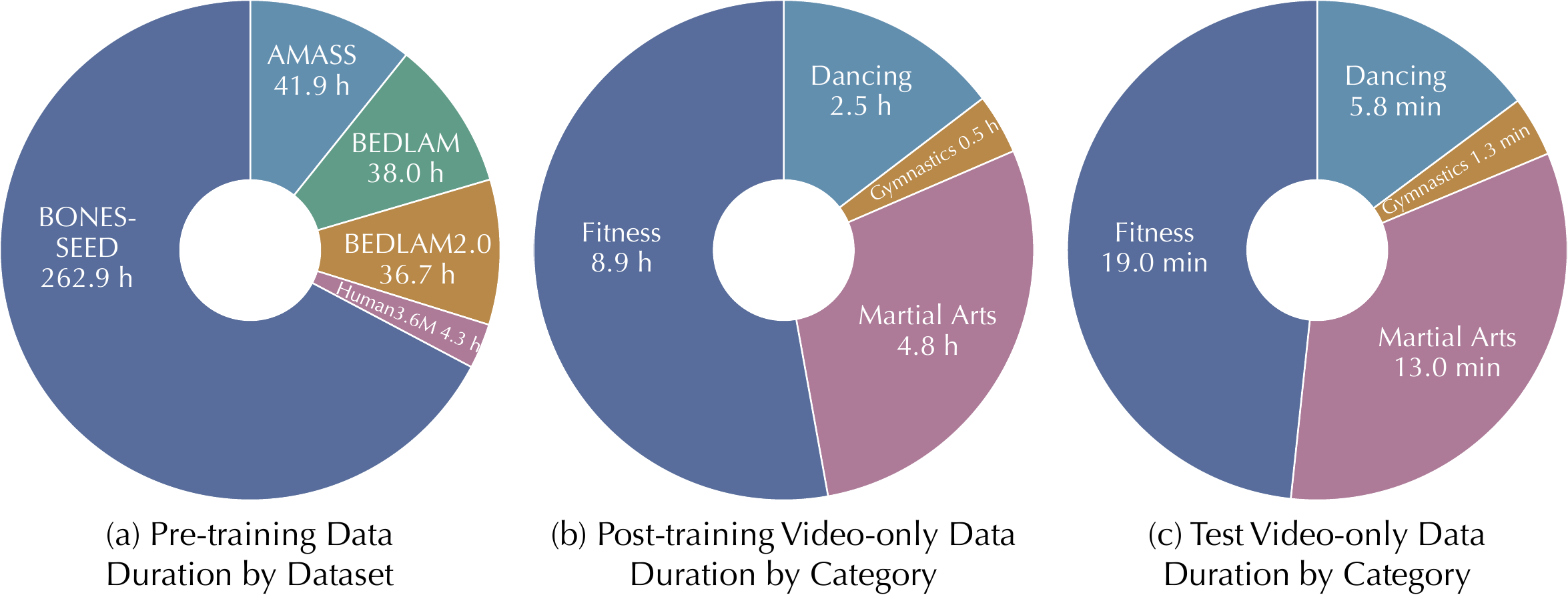}
    \caption{\textbf{Dataset statistics.} The charts show the duration of (a) each pre-training dataset, 383.8 hours in total, (b) each motion category in the video-only post-training data, 16.7 hours in total, and (c) each motion category in the held-out video-only test set, 39.1 minutes in total.}
    \label{fig:dataset}
\end{figure*}

\paragraph{Dataset statistics.} 
We summarize the dataset statistics for pre-training, post-training, and testing in \cref{fig:dataset}. For pre-training, we use 383.8 hours of motion data, of which 79 hours are paired with video input, while the remaining data are used without image features. For post-training, our video-only dataset consists of 8,000 videos totaling 16.7 hours and covers diverse motion categories, including dancing, fitness, gymnastics, and martial arts. Our held-out video-only test set contains 282 videos with a total duration of 39.1 minutes.

\paragraph{Adapted GMR retargeter.} 
For both training and testing datasets without paired Unitree G1 motions, we use GMR~\citep{araujo2025retargeting} to retarget the corresponding human motions to the G1 embodiment. We further adapt the original GMR pipeline to improve the robustness and quality of large-scale retargeting.

GMR performs frame-wise inverse kinematics and uses the optimized robot configuration from the previous frame to initialize the next frame. As a result, an inaccurate solution at the beginning of a sequence can propagate throughout the subsequent retargeting. To improve robustness, we run GMR multiple times with different initialization seeds. Each retargeted sequence is filtered and, when necessary, split according to the quality criteria described below, and we retain the result with the largest number of valid frames.

Following prior works on motion curation for humanoid retargeting~\citep{lee2026phuma,zhao2026make}, we design a set of rules to filter the source human motions and retargeted robot motions. For human motions, we filter sequences with excessive root jerk or insufficient foot contact. For robot motions, we reject sequences with a minimum root height below $0.25$ m, joint velocities exceeding the corresponding hardware limits, severe self-intersection, or excessive foot floating. Specifically, a frame is considered self-intersecting when the penetration depth exceeds $2$ cm, and we reject sequences in which such frames account for more than 5\% of the motion; we also require the mean lowest-foot height to remain below $0.10$ m.

For the test set, we additionally follow PHUMA~\citep{lee2026phuma} and apply a base-of-support criterion to the human motion. We construct the support region as the convex hull of the horizontal projections of the left and right ankle and foot joints, and require the average horizontal distances of the pelvis and spine1 from this region to be below $6$ cm and $11$ cm, respectively.

For sequences containing only localized invalid portions, we split the sequence into shorter segments and retain valid segments with at least 100 frames. Consequently, our adapted GMR retargeter does not necessarily preserve the length of the input sequence, but instead returns the candidate that retains the largest number of valid frames.

\paragraph{Data conversion between human representations.} Our humanoid reference motion generator takes human keypoints as input and does not rely on supervision from human body parameters. However, adapting the baseline methods to our training data requires a unified parametric representation of the human body. BONES-SEED represents human motion using SOMA~\citep{saito2026soma}, whereas all other datasets use SMPL-X~\citep{pavlakos2019expressive}. We therefore convert the human motions in BONES-SEED to the SMPL-X representation.

We first fit the SMPL-X shape parameters $\boldsymbol{\beta}\in\mathbb{R}^{10}$ for each actor. We reconstruct the actor-specific SOMA body mesh from its identity, scale, and bone-length parameters. Since the SOMA and SMPL-X models use different rest skeletons and mesh topologies, we re-pose the SOMA mesh using the rest-pose joint configuration of the corresponding SMPL-X model and then map the resulting mesh to the native SMPL-X topology. We then fit $\boldsymbol{\beta}$ to the transferred vertices, using $\ell_2$ regularization on the shape parameters and Huber weighting to reduce the influence of large vertex errors. With the fitted $\boldsymbol{\beta}$ fixed, we fit the remaining SMPL-X parameters using the official fitting script provided by SOMA.

\subsection{Human Study}\label{sec:human-study-detail}

We recruit 20 volunteer participants for the human study, all of whom have research backgrounds in computer vision or robotics, including 12 with prior experience in humanoid research. In each trial, participants are shown the original human demonstration video on top and two side-by-side humanoid motion results below. One result is produced by our method and the other by a baseline, with the left-right order randomized. Participants are asked to make a forced-choice judgment on which result better transfers the human motion depicted in the input video to the Unitree G1 humanoid.

\subsection{Evaluation Metrics}\label{sec:metric-detail}

\paragraph{Metrics for motion quality.} 
To evaluate the quality of the generated humanoid reference motions, we report local MPJPE and PA-MPJPE on root-relative joint positions. We additionally adopt \acf{rte} and motion jitter (Jitter) from WHAM~\citep{shin2024wham}. Self-penetration (Pen.) follows the metric used in OmniTrack~\citep{li2026omnitrack} and is reported in \textperthousand. Details of the 2AFC perceptual study are provided in \cref{sec:human-study-detail}.

\paragraph{Metrics for motion executability.} 
To evaluate the executability of the generated humanoid reference motions, we adopt the standard metrics used in humanoid motion tracking evaluation. The definition of tracking success is provided in \cref{sec:post-train-detail}. For EMDB~\citep{kaufmann2023emdb} and RICH~\citep{huang2022capturing}, we evaluate tracking success independently on each 250-frame segment and report the average across segments, whereas no fixed segment-length constraint is imposed for the video-only test sets. All remaining metrics are computed on root-relative joint positions. \Ac{mpjpe} and \ac{pampjpe} are computed between the tracked motion and the retargeted ground-truth motion, while Trk. \ac{mpjpe} and Trk. \ac{pampjpe} are computed between the reference motion and its tracked motion.

\subsection{Adaptation of Baselines}\label{sec:baseline-detail}

\paragraph{Adaptations.} Compared with the original GVHMR~\citep{shen2024world} and GEM~\citep{li2025genmo}, the adapted variants ($\dagger$) differ in training data, model capacity, and visual encoder. First, we train all baselines on the same datasets used to train our reference motion generator, with the data conversion between different representations described in \cref{sec:dataset-and-data-conversion-detail}. We set the number of Transformer layers, hidden dimension, and number of attention heads to be the same as those used in our method (\cref{sec:network-architecture-detail}). Finally, we replace the original HMR\,2.0 visual encoder~\citep{goel2023humans} with SAM 3D Body~\citep{yang2026sam3dbody,yang2026fastsam3dbody}, using the same image features as our method.

\paragraph{Training setup.} We train all adapted baselines on 4 NVIDIA H100 GPUs. For each baseline, we follow its original training hyperparameters while keeping the total training budget consistent with our method, using 730k optimization steps.

\begin{table*}[t!]
    \centering
    \small
    \caption{\textbf{Quantitative comparison of motion quality.} Results use ground-truth camera motion. $\dagger$~marks baselines retrained on our training data with a model size comparable to \model, and w/o PT denotes a generator before post-training. Lower is better for all metrics. Bold and underline mark the best and second-best results.}
    \label{tbl:comparison-gt-cam}
    \vspace{-0.25cm}
    \setlength{\tabcolsep}{9pt}
    \resizebox{\linewidth}{!}{%
        \begin{tabular}{lcccccccccc}
            \toprule
            \multirow{3}{*}{Method} & \multicolumn{5}{c}{EMDB} & \multicolumn{5}{c}{RICH} \\ 
            \cmidrule(r){2-6} \cmidrule(l){7-11} & \multirow{2}{*}{MPJPE} & PA- & \multirow{2}{*}{RTE} & \multirow{2}{*}{Jitter} & \multirow{2}{*}{Pen.} & \multirow{2}{*}{MPJPE} & PA- & \multirow{2}{*}{RTE} & \multirow{2}{*}{Jitter} & \multirow{2}{*}{Pen.} \\
            & & MPJPE & & & & & MPJPE & & & \\
            \midrule
            GVHMR + GMR & 73.3 & 54.4 & \underline{2.3} & 17.7 & 2.3 & 51.6 & 44.9 & 4.6 & \underline{7.7} & 1.7 \\
            GVHMR$^\dagger$ + GMR & 71.4 & 49.1 & 3.4 & 15.6 & 1.6 & 69.6 & 51.5 & \underline{4.4} & 10.3 & 2.2 \\
            GEM + GMR & 62.1 & 38.3 & 3.1 & 18.5 & 1.2 & 56.7 & 49.6 & 4.7 & 9.5 & 2.2 \\
            GEM$^\dagger$ + GMR & 68.8 & 46.1 & 3.3 & 20.5 & 1.3 & 70.3 & 51.2 & \underline{4.4} & 20.5 & 2.2 \\
            \midrule
            Ours (online) w/o PT & 55.4 & 36.6 & 8.3 & 24.0 & \underline{0.2} & 44.5 & 34.5 & 6.8 & 24.4 & 0.3 \\
            Ours (online) & 55.4 & 36.8 & 6.4 & 17.5 & \textbf{0.1}& 44.3 & 36.1 & 6.3 & 17.4 & \underline{0.2} \\
            \midrule
            Ours (offline) w/o PT & \textbf{52.9} & \textbf{34.0} & \textbf{1.6} & \underline{9.3} & \underline{0.2} & \underline{38.7} & \textbf{28.5} & \underline{4.4} & \textbf{7.2} & \underline{0.2} \\
            Ours (offline) & \underline{53.2} & \underline{34.2} & \textbf{1.6} & \textbf{8.9} & \textbf{0.1} & \textbf{38.4} & \underline{29.4} & \textbf{4.3} & \textbf{7.2} & \textbf{0.1} \\
            \bottomrule
        \end{tabular}
    }
\end{table*}

\begin{table*}[t!]
    \centering
    \small
    \caption{\textbf{Quantitative comparison of motion executability.} All reference motions are executed by $\pi_\text{track}$ in MuJoCo, using ground-truth camera motion. GT + GMR executes the retargeted ground-truth motions, and rows without GMR track the reconstructed human motions directly. $\dagger$ and w/o PT follow \cref{tbl:comparison-gt-cam}. Higher Succ.\ and lower errors are better. Bold and underline mark the best and second-best results, excluding GT + GMR.}
    \label{tbl:comparison-track-gt-cam}
     \vspace{-0.25cm}
    \setlength{\tabcolsep}{6pt}
    \resizebox{\linewidth}{!}{%
        \begin{tabular}{lcccccccccc}
            \toprule
            \multirow{3}{*}{Method} & \multicolumn{5}{c}{EMDB} & \multicolumn{5}{c}{RICH} \\ 
            \cmidrule(r){2-6} \cmidrule(l){7-11} & \multirow{2}{*}{Succ.} & \multirow{2}{*}{MPJPE} & PA- & Trk. & Trk. PA- & \multirow{2}{*}{Succ.} & \multirow{2}{*}{MPJPE} & PA- & Trk. & Trk. PA- \\
            & & & MPJPE & MPJPE & MPJPE & & & MPJPE & MPJPE & MPJPE \\
            \midrule
            GT + GMR & 91.0 & 35.2 & 22.7 & 35.2 & 22.7 & 100.0 & 30.3 & 19.6 & 30.3 & 19.6 \\
            \midrule
            GVHMR + GMR & 61.7 & 96.0 & 60.0 & 47.2 & 29.7 & 90.3 & \underline{64.7} & 47.2 & 36.4 & 29.0 \\
            GVHMR$^\dagger$ + GMR & 71.4 & 93.3 & 57.2 & 44.3 & 29.8 & 84.7 & 90.5 & 58.2 & 52.5 & 34.4 \\
            GEM + GMR & 74.4 & 84.3 & 41.5 & 39.0 & 23.9 & 90.3 & 71.6 & 51.3 & 37.8 & 29.5 \\
            GEM$^\dagger$ + GMR & 75.1 & 89.2 & 52.4 & 41.6 & 27.6 & 90.3 & 88.8 & 55.1 & 50.8 & 33.4 \\
            GVHMR & 54.9 & 87.5 & 46.0 & - & - & 29.2 & 83.6 & 46.4 & - & - \\
            GVHMR$^\dagger$ & 56.4 & 84.0 & 45.8 & - & - & 26.4 & 82.0 & 46.5 & - & - \\
            GEM & 54.9 & 85.4 & 45.0 & - & - & 30.6 & 86.7 & 47.7 & - & - \\
            GEM$^\dagger$ & 56.4 & 83.4 & 45.4 & - & - & 23.6 & 82.0 & 46.2 & - & - \\
            \midrule
            Ours (online) w/o PT & 76.2 & 82.2 & 41.3 & 31.7 & 22.4 & 87.5 & 76.8 & 41.5 & 35.0 & 22.2 \\
            Ours (online) & 87.7 & 83.1 & 40.3 & \textbf{29.4} & \textbf{16.4} & \underline{98.2} & 69.8 & 40.1 & 27.7 & \textbf{17.3} \\
            \midrule
            Ours (offline) w/o PT & \underline{89.5} & \underline{73.9} & \underline{40.1} & 32.0 & 18.6 & \textbf{100.0} & \textbf{62.0} & \textbf{34.5} & \underline{26.2} & 18.0 \\
            Ours (offline) & \textbf{90.2} & \textbf{71.8} & \textbf{39.7} & \underline{31.3} & \underline{18.0} & \textbf{100.0} & \textbf{62.0} & \underline{34.6} & \textbf{25.9} & \underline{17.7} \\
            \bottomrule
        \end{tabular}
    }
\end{table*}

\section{Additional Experimental Results}\label{sec:experiments-results}

In this section, we provide additional experimental results. We report quantitative comparisons with the baselines and ablation variants using ground-truth cameras (\cref{sec:comparison-ground-truth-camera}). We provide qualitative comparisons against the baselines and ablation variants on our \href{https://awfuact.github.io/vit-h/}{\textit{project page}}.

\subsection{Comparisons with Ground-Truth Camera}\label{sec:comparison-ground-truth-camera}

\Cref{tbl:comparison-gt-cam,tbl:comparison-track-gt-cam} repeat the comparisons of \cref{tbl:comparison,tbl:comparison-track} on EMDB and RICH with the ground-truth cameras provided by the datasets. The trends match those with estimated cameras. The offline generator outperforms all pipelines on every metric in both tables, and both post-trained generators achieve higher success rates and lower tracking errors than all pipelines. Post-training reduces self-penetration in every setting and jitter in most, lowers tracking errors, and raises or preserves success rates, while keeping local accuracy comparable.

\section{Limitations and Future Work}\label{sec:limitations-and-future-work}

\paragraph{Limitations.}
Our experiments demonstrate favorable scaling behavior across the model and data scales considered in this work, for both pre-training and post-training. However, the extent of our scaling study is limited by the available training data and computational resources, and we therefore do not explore substantially larger models or datasets beyond the regime studied here. Additionally, our current motion representation and tracking setup do not model dexterous hand poses. Consequently, the object-interaction behaviors considered in this work are primarily limited to whole-body manipulation of relatively large and simple objects, rather than fine-grained manipulation such as grasping small objects.

\paragraph{Future work.}
A natural next step is to further scale \model with larger models, more diverse pre-training and post-training data, and increased computational resources. Beyond DPO~\citep{rafailov2023direct}, it would also be interesting to explore alternative post-training algorithms, such as PPO~\citep{schulman2017proximal} or GRPO~\citep{shao2024deepseekmath}, for directly optimizing motion quality and executability. Moreover, while our experiments show that the improvements from motion-tracking feedback generalize across different trackers, future work could incorporate feedback from multiple trackers during post-training to provide a more robust executability signal. Finally, extending the motion representation and tracking framework to model hand poses would enable finer-grained object interactions, including grasping and manipulation of smaller objects.

\end{document}